\documentclass[10pt,journal,compsoc]{IEEEtran}
\ifCLASSOPTIONcompsoc
  \usepackage[nocompress]{cite}
\else
  \usepackage{cite}
\fi
\ifCLASSINFOpdf
\else
\fi

\usepackage[ruled]{algorithm2e}
\usepackage[utf8]{inputenc} 
\usepackage[T1]{fontenc}    
\usepackage{hyperref}       
\usepackage{url}            
\usepackage{booktabs}       
\usepackage{amsfonts}       
\usepackage{nicefrac}       
\usepackage{microtype}      
\usepackage{xcolor}         
\usepackage{amsmath}
\usepackage{amssymb}
\usepackage{multirow}
\usepackage{graphicx}
\usepackage{subfigure}
\usepackage{algorithmic}
\usepackage{makecell}
\usepackage{wrapfig}
\usepackage{lipsum}
\def\A{{\bf A}}

\def\D{{\bf D}}

\def\H{{\bf H}}
\def\I{{\bf I}}

\def\X{{\bf X}}

\def\Z{{\bf Z}}

\def\y{{\bf y}}
\def\z{{\bf z}}

\def\0{{\bf 0}}
\def\1{{\bf 1}}

\def\IM{{\mathcal I}}
\def\GM{{\mathcal G}}

\def\LM{{\mathcal L}}

\def\DM{{\mathcal D}}

\def\VM{{\mathcal V}}
\def\EM{{\mathcal E}}

\def\ie{{\em i.e.}}

\begin{document}
%
\title{Deep Hypergraph Structure Learning}
%
%
%
%

\author{Zizhao~Zhang, Yifan~Feng, Shihui~Ying,~\IEEEmembership{Member,~IEEE,}
        and Yue~Gao,~\IEEEmembership{Senior Member,~IEEE,}
\IEEEcompsocitemizethanks{\IEEEcompsocthanksitem Z. Zhang, Y. Feng and Y. Gao are with the School of Software, Tsinghua University, Beijing 100084, China. 

E-mail: zhangziz18@mails.tsinghua.edu.cn; evanfeng97@gmail.com; kevin.gaoy@gmail.com. \protect\\
\IEEEcompsocthanksitem S. Ying is with the Department of Mathematics, School of Science, Shanghai University, Shanghai 200444, China.

E-mail: shying@shu.edu.cn.}
}

%
%

\markboth{IEEE TRANSACTIONS ON PATTERN ANALYSIS AND MACHINE INTELLIGENCE}%
{Shell \MakeLowercase{\textit{et al.}}: Bare Demo of IEEEtran.cls for Computer Society Journals}
%



\IEEEtitleabstractindextext{%
\begin{abstract}
Learning on high-order correlation has shown superiority in data representation learning, where hypergraph has been widely used in recent decades. The performance of hypergraph-based representation learning methods, such as hypergraph neural networks, highly depends on the quality of the hypergraph structure. How to generate the hypergraph structure among data is still a challenging task. Missing and noisy data may lead to "bad connections" in the hypergraph structure and destroy the hypergraph-based representation learning process. Therefore, revealing the high-order structure, \ie, the hypergraph behind the observed data, becomes an urgent but important task. To address this issue, we design a general paradigm of deep hypergraph structure learning, namely DeepHGSL, to optimize the hypergraph structure for hypergraph-based representation learning.  
Concretely, inspired by the information bottleneck principle for the robustness issue, we first extend it to the hypergraph case, named by the hypergraph information bottleneck (HIB) principle. Then, we apply this  principle to guide the hypergraph structure learning, where the HIB is introduced to construct the loss function to minimize the noisy information in the hypergraph structure. The hypergraph structure can be optimized  and this process can be regarded as enhancing the correct connections and weakening the wrong connections in the training phase. Therefore, the proposed method benefits to extract more robust representations even on a heavily noisy structure.  Finally, we evaluate the model on four benchmark datasets for representation learning. The experimental results on both graph- and hypergraph-structured data demonstrate the effectiveness and robustness of our method compared with other state-of-the-art methods. 
\end{abstract}

}

\maketitle

\IEEEdisplaynontitleabstractindextext

%

%
\IEEEpeerreviewmaketitle

\section{Introduction}

High-order correlation widely exists in real applications like neuronal networks, social networks, and transportation networks \cite{benson2016higher}. In recent years, the value of high-order correlation has been intensively studied in representation learning \cite{feng2018hypergraph, wu2020adahgnn}. The hypergraph is a flexible and powerful mathematical tool to model the high-order correlation. The hypergraph-based representation learning, represented by hypergraph neural networks \cite{chien2022you, liao2021hypergraph}, has attracted much attention and been widely applied to many tasks, including drug discovery~\cite{shalini2018drugs}, 3D pose estimation~\cite{liu2020semi}, action recognition~\cite{hao2021hypergraph}, recommendation system~\cite{zheng2018novel, bu2010music}, collaborative networks~\cite{zhang2010hypergraph}, etc.  
Hypergraph neural networks are a series of models that convolute node signals on a hypergraph structure and output predictions for downstream tasks. A hypergraph is composed of a set of nodes and a set of hyperedges, in which each hyperedge can connect more than two nodes. Compared with the simple graph, the hypergraph structure is more powerful and flexible to model high-order complex correlations among data. Hypergraph neural networks embed the high-order correlation into node representation and thus exhibit their superior performance.

\begin{figure}[] 
	\centering 
	\includegraphics[width=\linewidth]{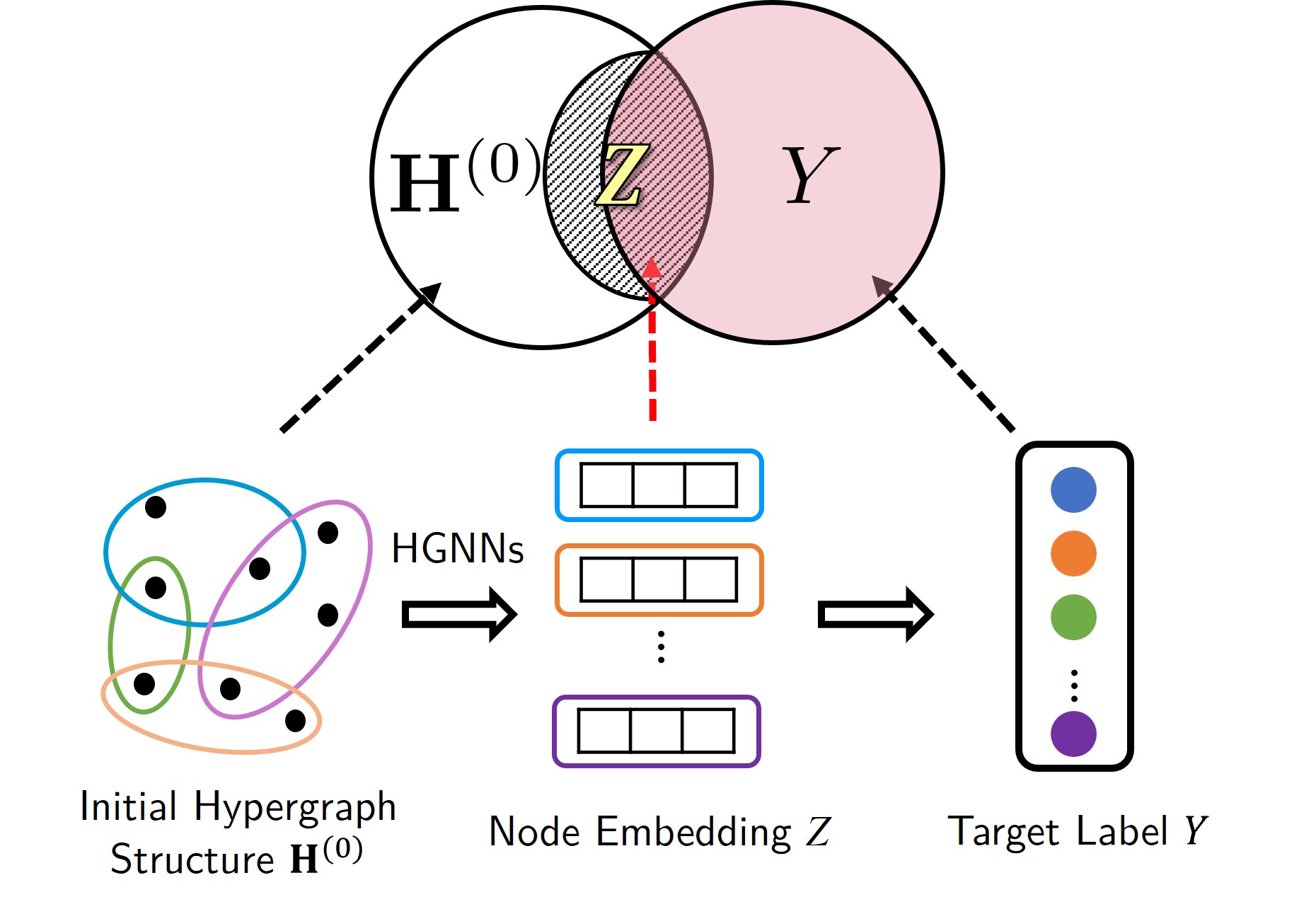} 
	\caption{An illustration of hypergraph information bottleneck. In the information of node embedding $Z$ (shadow area in the middle), the mutual information $\IM(Z,Y)$ (pink shadow area) is useful for the downstream task and the extra part of mutual information $\IM(Z,\H^{(0)})$ which is unrelated to $Y$ (white shadow area) is noise.} 
	\label{Fig.hib} 
\end{figure}

However, existing hypergraph neural networks all hold the promise that nodes connected by the same hyperedge should have similar representations, which causes the performance of representation learning to overly rely on the hypergraph structure. 
When the hypergraph structure contains a considerable amount of noisy information, such noisy information  will be integrated into the learned node representations, which may lead to poor performance of hypergraph neural networks. For example, nodes with different labels may belong to the same hyperedge and decrease the discrimination of data in the classification task. What's more, there exist difficulties to obtain the high-order connections among data in some cases, discovering the high-order correlations among multiple brain regions in the brain networks for instance. The missing connections may lead to an incomplete hypergraph structure and destroy hypergraph neural networks. 
We name the noisy and missing connections in the hypergraph structure as "bad connections", which have a negative influence on the performance of hypergraph neural networks on downstream tasks. To overcome this issue, it is essential to optimize the  hypergraph structure by deleting or decreasing the weights of task-irrelevant connections, and adding or strengthening the connections that significantly contribute to the downstream tasks. The optimal hypergraph structure is expected to formulate data high-order correlations more accurately and completely than the initial one.

The information bottleneck (IB) is a widely used principle to improve the robustness of the representation learning to noisy data \cite{tishby2000information, alemi2016deep, saxe2019information}, which aims to preserve the mutual information of the learned representation $Z$ and the target $Y$ as much as possible while minimizing the mutual information of $Z$ and the input feature $X$, namely the minimum sufficient principle. This principle limits $Z$ to contain too much noisy information about $X$ but unrelated to $Y$. Therefore, to overcome the bad connections in the hypergraph structure, a natural idea is to explore the hypergraph information bottleneck principle. The aim of hypergraph information bottleneck (HIB) is to encourage the learned hypergraph representation to be sufficiently informed about the prediction task, and conversely discourage the representation from involving redundant information about the initial hypergraph structure, as illustrated in Fig. \ref{Fig.hib}. In mathematical, the HIB principle is to minimize the mutual information of $\IM(Z,\H^{(0)})$ (minimum), while maximizing the mutual information of $\IM(Z,Y)$ (sufficient).

There have been some methods applying IB to representation learning on unstructured data \cite{tishby2015deep} and graph-structured data~\cite{wu2020graph, yu2020graph, yang2021heterogeneous, yu2022improving}. The IB principle can effectively make the learning model more robust and shows superior performance, especially in tasks like adversarial attacks. However, these methods can not be directly extended and applied to hypergraph-structured data for two reasons. First, the information bottleneck of hypergraph learning depends on not only the features but also the initial structure $\H^{(0)}$. It is intractable to instantiate the mutual information  $\IM(\H^{(0)};Z)$ due to the complex dependencies among these nodes and minimize it in a trainable way. Second, 
the information aggregation in current hypergraph neural networks is based on the fixed structure, leading to the information related to the initial hypergraph in the learned representation being unbending. It is a necessity to learn the hypergraph structure adaptively.

To solve the noisy hypergraph structure problem and bring HIB to real applications, we propose a deep hypergraph structure learning framework, DeepHGSL, to optimize the hypergraph structure for robust representation learning.  We present a general paradigm in which the hypergraph structure evolves in an iterative manner. This paradigm can be integrated with hypergraph neural networks. We further propose the hypergraph information bottleneck principle and give an estimation of HIB to guide the modification of the hypergraph structure.  HIB is instantiated as a linearly scalable and gradient backpropagable loss function in our framework. In our experiments, the general paradigm is implemented based on the spatial hypergraph convolution and hyperedge-vertex attention mechanism. A benchmark dataset for representation learning tasks on hypergraph, Yummly10k, is made to evaluate the effectiveness and robustness of DeepHGSL. A comparison study with the cutting-edge graph/hypergraph representation learning methods on both graph-structured data and hypergraph-structured data demonstrates that our method can achieve robust performance confronting the missing or noisy high-order connections existing in the initial hypergraph.

The contributions of this paper are mainly three folds. 
\begin{itemize}
    \item Firstly, we design a deep hypergraph structure learning
    paradigm, DeepHGSL. Here, we propose a hypergraph structure learning strategy and integrate it with hypergraph neural networks for representation learning. The hypergraph structure is iteratively optimized to formulate the high-order correlation among data more accurately, and thus allow the representation learning more robust.
    \item  Secondly, we propose the hypergraph information bottleneck and instantiate it by estimating the variational lower and upper bounds of the mutual information. The estimation is further used as the loss function to guide deep hypergraph structure learning. 
    \item  Thirdly, we give an implementation of the whole framework and prove the convergence property of the algorithm. We evaluate the performance of DeepHGSL on two graph datasets and two hypergraph datasets.  We made a new hypergraph-structured dataset, Yummly10k, in which the high-order correlations among data are natural and inherent. The experimental results have demonstrated the superiority and robustness of our proposed method, especially on the hypergraph-structured data.
\end{itemize}
 
The rest of this paper is organized as follows. We first introduce the related work and preliminary in Sections \ref{sc:relatedwork} and \ref{sc:preliminary}. We then provide a general paradigm of DeepHGSL in Section \ref{sc:paradigm}. The hypergraph information bottleneck is analyzed in detail in Section \ref{HIB}. An implementation of the DeepHGSL framework is given in Section \ref{sc:implementation}. Experiments and discussions are further provided in Section \ref{sc:experiement}. 

\section{Related Work} \label{sc:relatedwork}

\subsection{Hypergraph Modeling and Hypergraph Structure Learning}
A traditional pipeline for hypergraph learning framework is firstly constructing a hypergraph structure to formulate the complex correlation among data, and then learning on the initial hypergraph structure to conduct downstream tasks. The hypergraph construction methods can be categorized into explicit  and implicit depending on the input data. If there are explicit high-order correlations among data, we can directly build hyperedges according to such correlations. For example, in recommendation systems, we can construct a hypergraph by user-item relationships in which the users are nodes and each item is represented as a hyperedge \cite{ji2020dual}. In contrast, we need to implicitly build hypergraphs when such high-order correlations are unavailable. In \cite{zhang2018inductive}, the features of 3D objects are used to build a $k$-NN hypergraph, each node is connected with its $k$ nearest neighborhoods. The $l1$ sparsity representation of nodes is learned to build hyperedges in \cite{wang2015visual}.  Although the above methods can conduct hypergraph modeling using various input data, the structure is fixed after construction. 

To be more adaptive to different data and tasks, Gao et al. \cite{gao2012visual} assigned different weights to hyperedges and adaptive adjust the weights during the learning process. In \cite{su2017vertex}, the vertices are further weighted to estimate their importance. For multimodal data, each modality establishes a corresponding sub-hypergraph structure, and these sub-hypergraphs are weighted and fused by learnable weights in \cite{zhang2018inductive}. Since the hypergraph is a higher-order correlation structure, we can learn not only the weights of hyperedges, nodes, and sub-hypergraphs but also the hyperedge-dependent vertex weights like scoring the possibility of some vertex belonging to some hyperedge. To achieve this, Zhang et al. \cite{zhang2018dynamic} directly optimized the incidence matrix of hypergraph via gredient descending. Although there have been many methods for hypergraph structure optimization in traditional hypergraph learning, there is still a lack of hypergraph structure optimization methods in deep hypergraph learning.

\subsection{Hypergraph Neural Networks}
\begin{figure*}[] 
	\centering 
	\includegraphics[width=0.8\linewidth]{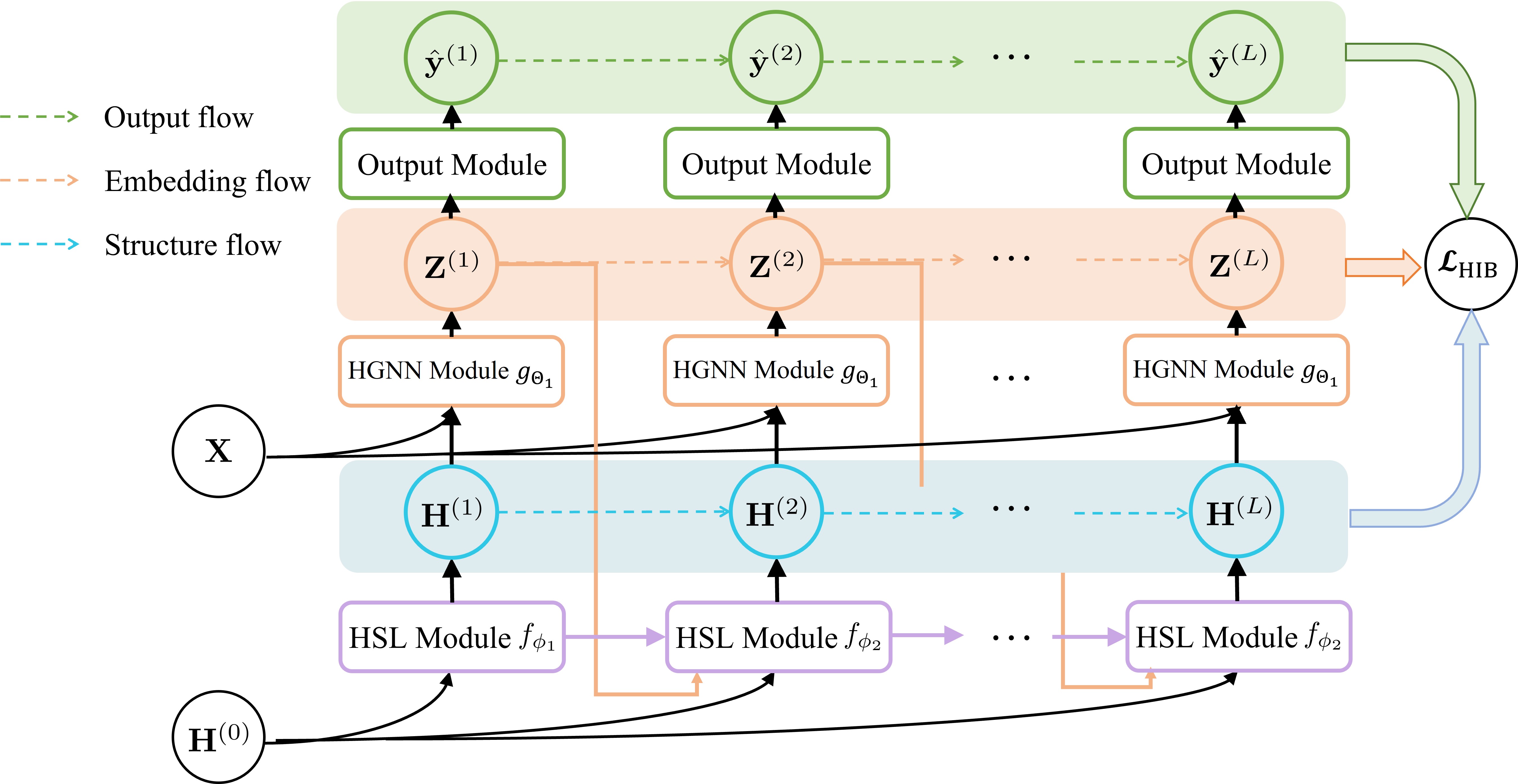} 
	\caption{Pipeline of the deep hypergraph structure learning paradigm.} 
	\label{Fig.pipeline} 
\end{figure*}
There is a body of existing work on hypergraph neural networks for analyzing hypergraph-structured data. Compared with graph modeling, hypergraph modeling is more flexible to describe the complex data correlation (connections involving three nodes or more), and have attracted increasing attention \cite{gao2020hypergraph} in recent years. The deep hypergraph representation learning methods introduce hypergraph modeling into a more complex deep representation learning framework, which can be broadly categorized into three classes: spectral-based, spatial-based, and diffusion-based. The representative work of spectral-based methods is hypergraph neural networks \cite{feng2018hypergraph} and HyperGCN \cite{yadati2019hypergcn}. Hypergraph neural networks \cite{feng2018hypergraph} first formulate the spectral convolution on hypergraph structure as a neural network layer, which uses the hypergraph Laplacian eigenbasis to approximate the Fourier transform. In HyperGCN \cite{yadati2019hypergcn}, each hyperedge is estimated as a simple graph by selecting two central vertices and connecting the rest nodes in the hyperedge with these two centers, and thus the original problem is approximated as a graph learning problem. 

Different from the spectral-based methods which define the hypergraph convolutions from the hypergraph spectral theory, spatial-based methods formulate hypergraph convolution as aggregating information from local neighborhoods. HyperSAGE \cite{arya2020hypersage} characterized the spacial-based message passing process on hypergraph by a two-stage procedure, \ie, from nodes to hyperedges $v\rightarrow e$ and from hyperedges to nodes $e\rightarrow v$. To solve the high computational cost and over-smoothing issue in HyperSAGE, UniGNN \cite{huang2021unignn} further proposed a general framework to describe the spacial-based message propagation method in both graph and hypergraph neural networks. Lastly, the insight of diffusion-based methods is that the message passing process can be an analogy to spreading labels on the hypergraph. The main advantage of diffusion-based methods is that they work typically fast. In the diffusion method HyperND \cite{prokopchik2022nonlinear}, the node embedding  is repeatedly updated by a nonlinear diffusion map until achieving the limit point. Then, the final embedding is fed into a fully-connected (FC) layer for prediction. We note that the diffusion module and FC layer in HyperND are decoupled and only parameters in the FC layer need training. Therefore, HyperND is much faster than hypergraph neural networks.

\section{Preliminary of Hypergraph and Hypergraph Neural Networks} \label{sc:preliminary}
Formally, a hypergraph is defined by a set of nodes $\VM$ and a set of hyperedges $\EM$. Each hyperedge denotes an interaction in which one or more nodes can participate. We can assign different weights to different components in the hypergraph, including the hyperedges, vertices, and hyperedge-vertex correlations. 
The hyperedge-dependent vertex weights not only denote whether a node belongs to a hyperedge (0 or 1), but also represent the importance of different nodes in the same hyperedges or the probabilities of affiliation relationships between nodes and hyperedges (in the range of $[0,1]$). The hyperedge-dependent vertex weights are represented by an incidence matrix $\H \in \mathbb{R}^{|\VM|\times|\EM|}$. Since there is no constraint on the number of nodes a hyperedge can connect, the hypergraph provides a general and flexible description of high-order interactions. 

In a hypergraph with hyperedge-dependent vertex weights, the degree of vertex $v$ is defined by $d(v)=\sum_{e \in \mathcal{E}} \mathbf{H}(v, e)$, and the degree of hyperedge $e$ is defined by $d(e)=\sum_{v \in \mathcal{V}} \mathbf{H}(v, e)$. Let $\D_v$ and $\D_e$ denote the diagonal matrices whose diagonal items are the vertices degrees and hyperedge degrees, respectively. According to \cite{zhou2006learning} and \cite{feng2018hypergraph}, the laplacian matrix of the hypergraph is defined as
\begin{equation}
	\Delta = \I - \D_v^{-1/2}\H\D_e^{-1}\H^\top\D_v^{-1/2},
\end{equation}
and the hypergraph convolution on the spectral domain is parameterized as:
\begin{equation}
	\X^{t+1}  = \D_v^{-1/2}\H\D_e^{-1}\H^\top\D_v^{-1/2}\X^{t}\mathbf{\Theta} 
\end{equation}
where $\mathbf{\Theta} \in \mathbb{R}^{d_{in}\times d_{out}}$ is the trainable parameters in the hypergraph convolution layer. In \cite{gao2022hgnn}, motivated by hyper-path in the hypergraph, the spatial-based convolution on hypergraphs named HGNNConv$^+$ can be written in the matrix format as
\begin{equation}
\X^{t+1}  = \D_v^{-1}\H\D_e^{-1}\H^\top\X^{t}\mathbf{\Theta},
\end{equation}
where $\mathbf{\Theta} \in \mathbb{R}^{d_{in}\times d_{out}}$ is also the trainable parameter. In the first stage of HGNNConv$^+$, the vertex features are transformed into the hyperedge features by the incidence matrix $\mathbf{H}^\top$, and the hyperedge features are further transformed into new vertex features following the incidence matrix $\mathbf{H}$. The spectral-based convolution performs the vertex to vertex feature smoothing with the pre-defined hypergraph Laplacian matrix. However, the spatial-based convolution performs a two-stage message passing of vertex-hyperedge-vertex, which is more flexible and can be easily extended to more types of high-order structures like directed hypergraphs.

\label{paradigm}
\section{A General Paradigm of DeepHGSL} \label{sc:paradigm}

\subsection{Problem Formulation} 
The hypergraph-structured data includes a hypergraph structure $\GM=(\VM,\EM)$ and the node features $\X$. The set $\VM$ contains $n$ nodes and the set $\EM$ contains $m$ hyperedges. Let $\H^{(0)}$ denotes the initial noisy hyperedge-dependent vertex weight. Given a noisy hypergraph input $\DM:=\{\H^{(0)},\X\}$, the problem we focus on in this work is to obtain an optimized hypergraph structure $\H^{(*)}$ and extract its corresponding node-level representations $\Z = f(\H^{(*)},\theta)$ simultaneously. The node embedding can be further used for the downstream node classification task where nodes are associated with some labels $Y \in[K]^{n}$.

\subsection{Pipeline} 
The whole framework contains two kinds of modules: hypergraph structure learning modules (HSL modules) for optimizing the hypergraph structure and hypergraph neural network modules (HGNN modules) for node representation learning. These two kinds of modules are stacked alternatively to form the whole framework. The whole framework totally contains $L$ layers and each layer is composed of one HSL module, one HGNN module, and one output module. Specifically, the $l$-th layer HSL module establishes a new hypergraph structure based on the node representation of the previous layer and combines it with the initial hypergraph structure. Then, the HGNN module re-convolves the original signal on the updated hypergraph structure. 

Note that the first HSL module takes raw feature vectors as input, while other HSL modules take node embeddings as input, leading to different parameter dimensions. Therefore, the first HSL module has independent parameters and combines the new structure with the initial one, and later HSL modules share the same parameters, combining the output with not only the initial structure but also the one produced by the previous HSL module. The $l$-th layer HSL module is formulated as

\begin{align}
	\H^{(1)} &= f_{\Phi_1}(\X, \H^{(0)}); \notag \\  \H^{(l)} &= f_{\Phi_2}(\Z_{\VM}^{(l-1)}, \H^{(l-1)}, \H^{(0)}),
\end{align}
where $\Phi_1$ are the network parameters in the first HSL module, and $\Phi_2$ are the shared network parameters in later HSL modules.

All HGNN modules share network parameters. Denote the parameters in HGNN module as $\mathbf{\Theta}_{1}$, the node embedding of the $l$-th layer HGNN module $\Z_{\VM}^{(l)}$ is written as
\begin{equation}
	\Z_{\VM}^{(l)} = g_{\mathbf{\Theta}_{1}}(\H^{(l)}, \X) \label{eq:hid}
\end{equation}
Each layer has an output module that maps the node embedding to label prediction. All output modules share the same parameters, denoted as $\mathbf{\Theta}_{2}$. The output function of the $l$-th layer is written as
\begin{equation}
	\hat{\y}^{(l)} = g_{\mathbf{\Theta}_{2}}(\H^{(l)}, \Z_{\VM}^{(l)}) \label{eq:out}
\end{equation}
Finally, the output flow $\{\hat{\y}^{(1)}, \dots, \hat{\y}^{(L)}\}$ and the modified hypergraph structure flow $\{\H^{(1)}, \dots, \H^{(L)}\}$ are all used to calculate the loss function, which is designed based on the HIB principle as will be introduced in the following. 

\section{Hypergraph Information Bottleneck}
\label{HIB}
The hypergraph structure may exist noisy and missing connections and be prone to task-irrelevant connections. Embedding these bad connections will lead to poor performance of the downstream tasks. The aim of deep hypergraph structure learning is to encourage the hypergraph structure to evolve to model the correlation of data labels more accurately, while minimizing the additional information in the original hypergraph structure which is irrelevant to the target task. To balance the expressiveness and robustness of the model, we design the objective function based on the hypergraph information bottleneck principle, aiming to optimize the hypergraph structure to capture \textit{minimal sufficient} information for the downstream prediction task.
\subsection{HIB Principle}
As shown in the general paradigm, the original structure is adjusted in each iteration and forms the structure flow $\{\H^{(l)}\}_{1\leq l \leq L}$, which subsequently controls the following embedding flow and output flow. 
Based on the general paradigm, The HIB principle for the $l$-th layer embedding $\Z^{(l)}_{v}$ is written as
\begin{align}
	\min _{\mathbb{P}(\Z^{(l)}_{v} \mid \H^{(0)}, \X) \in \Omega} & \operatorname{HIB}_{\beta}(\H^{(0)}, Y ; \Z^{(l)}_{\VM}) \notag\\& \triangleq -\IM(Y ; \Z^{(l)}_{\VM})+\beta \IM(\H^{(0)} ; \Z^{(l)}_{\VM}),
	\label{eq:HIB}
\end{align}
where $\Omega$ declares the search space of the conditional distribution of $\Z^{(l)}_{v}$ given the initial structure $\H^{(0)}$ and data $\X$. The probabilistic dependence among $(\Z^{(l)}_{v}$, $\H^{(0)})$ and $\X$ is defined by the general paradigm. The trade-off hyperparameter $\beta$ is to balance the weights of two items.

\textbf{Proposition 1.} $\IM(\H^{(0)} ; \Z^{(l)}_{\VM}) \leq \IM(\H^{(0)} ; \H^{(l)})$.

\textbf{Proof.} Since the conditional distribution of $\Z^{(l)}_{\VM}$ depends only on $\H^{(l)}$ and is conditionally independent of $\H^{(0)}$, $\{\H^{(0)}, \H^{(l)}, \Z^{(l)}_{\VM}\}$ forms a Markov chain in this order, denoted as $\H^{(0)} \to \H^{(l)} \to \Z^{(l)}_{\VM}$. According to data-processing inequality, we have
\begin{equation}
	\IM(\H^{(0)} ; \Z^{(l)}_{\VM}) \leq \IM(\H^{(0)} ; \H^{(l)})
	\label{eq:upperbound1}
\end{equation}

Plugging Eq. (\ref{eq:upperbound1}) into the HIB principle Eq. (\ref{eq:HIB}), we obtain an upper bound as the objective function to minimize
\begin{equation}
	\LM_{\text{HIB}}^{(l)} = -\IM(Y ; \Z^{(l)}_{\VM})+\beta \IM(\H^{(0)} ; \H^{(l)}) \label{eq:obj}
\end{equation}

\subsection{Estimation of HIB}
To optimize the parameters of the model, we need to specify the mutual information $\IM(Y ; \Z^{(l)}_{\VM})$ and $\IM(\H^{(0)} ; \H^{(l)})$ in Eq. (\ref{eq:obj}). However, exact computing the mutual information is intractable if the probability distributions are unknown. Here we will explain how to estimate these two items to make them easily achieved, linearly scalable, and trainable through back-prop.

We first transform the problem of minimizing $-\IM(Y ; \Z^{(l)}_{\VM})$ into the problem of maximizing the lower bound of $\IM(Y ; \Z^{(l)}_{\VM})$. The Nguyen, Wainright \& Jordan’s bound $I_\text{NWJ}$ \cite{nguyen2010estimating} is used here:

\textbf{Lemma 1} \cite{nguyen2010estimating} For any two random variables $X_{1}, X_{2}$ and any bivariate function $g: g\left(X_{1}, X_{2}\right) \in \mathbb{R}$, we have
\begin{align}
	I\left(X_{1}, X_{2}\right) \notag\geq & \mathbb{E}\left[g\left(X_{1}, X_{2}\right)\right]\\&-\mathbb{E}_{\mathbb{P}\left(X_{1}\right) \mathbb{P}\left(X_{2}\right)}\left[\exp \left(g\left(X_{1}, X_{2}\right)-1\right)\right] \label{eq:lemma1}
\end{align}

The above lemma is used to $(Y ; \Z^{(l)}_{\VM})$. Plugging in $g\left(Y, \Z^{(l)}_{\VM}\right)=1+\log \frac{\prod_{v \in V} \operatorname{Cat}\left(\hat{\y}_v^{(l)}\right)}{\mathbb{P}(Y)}$ and ignoring constants, the right hand side of Eq. (\ref{eq:lemma1}) is substituted by the cross entropy loss, \ie,
\begin{equation}
	-\IM(Y ; \Z^{(l)}_{\VM}) \rightarrow \sum_{v \in V}  CE \left(\hat{\y}_v^{(l)} ; Y_{v}\right)
\end{equation}

The second term of Eq. (\ref{eq:obj}) $\IM(\H^{(0)} ; \H^{(l)})$ measures the mutual information between the initial hypergraph structure and the updated hypergraph structure. It has an upper bound and can be derived to a tractable objective to optimize as follows.

\textbf{Proposition 2} For any distribution $\mathbb{Q}(\H^{(l)})$ for $\H^{(l)}\in\{0,1\}^{|\VM|\times|\EM|}$, we have
\begin{equation}
	\IM(\H^{(0)} ; \H^{(l)}) \leq D_{K L}\left(\mathbb{P}(\H^{(l)} \mid \H^{(0)}) \| \mathbb{Q}(\H^{(l)})\right)
\end{equation}
\textbf{Proof} The mutual information between $\H^{(0)}$ and $\H^{(l)}$ is written as 
\begin{equation}
\IM(\H^{(0)} ; \H^{(l)})=\mathbb{E}_{\mathbb{P}(\H^{(0)} , \H^{(l)})} \left[\log \frac{\mathbb{P}(\H^{(l)} \mid \H^{(0)})}{\mathbb{P}(\H^{(l)})}\right] \notag
\end{equation}
The KL divergence distance between $\mathbb{Q}(\H^{(l)})$ and $\mathbb{P}(\H^{(l)})$ is no less than zero
\begin{align}
	&D_{K L}(\mathbb{P}(\H^{(l)}), \mathbb{Q}(\H^{(l)}))\notag\\ & =\mathbb{E}_{\mathbb{P}(\H^{(l)})}\left[\log \mathbb{P}(\H^{(l)})\right] -\mathbb{E}_{\mathbb{P}(\H^{(l)})}\left[\log \mathbb{Q}(\H^{(l)})\right] \notag\\& \geq 0 \notag
\end{align}
Thus, we have
\begin{align}
	\IM(\H^{(0)} ; \H^{(l)}) & \leq \mathbb{E}_{\mathbb{P}(\H^{(0)} , \H^{(l)})} \left[\log \frac{\mathbb{P}(\H^{(l)} \mid \H^{(0)})}{\mathbb{Q}(\H^{(l)})}\right] \notag
	\\& \approx \mathbb{E}_{\mathbb{P}(\H^{(0)} \mid \H^{(l)})} \left[\log \frac{\mathbb{P}(\H^{(l)} \mid \H^{(0)})}{\mathbb{Q}(\H^{(l)})}\right] \notag \\
	& = D_{K L}\left(\mathbb{P}(\H^{(l)} \mid \H^{(0)}) \| \mathbb{Q}(\H^{(l)})\right) \notag
\end{align}

To specify the upper bound for $\IM(\H^{(0)} ; \H^{(l)})$, we assume $\mathbb{Q}(\H^{(l)})$ is a non-informative prior and the elements in $\H^{(l)}$ are i.i.d Bernoulli distributions: $\H^{(l)} = \cup_{i,j}\left\{h_{ij} \in \{0,1\} \mid h_{ij} \stackrel{\mathrm{iid}}{\sim} \operatorname{Bernoulli}(0.5)\right\}$. We assume that elements in $\mathbb{Q}(\H^{(l)})$ have a probability of 0.5 to be 1 or 0 because there is no prior information about whether vertex $v$ belongs to hyperedge $e$ or not. Thus, the estimation of $\IM(\H^{(0)} ; \H^{(l)})$ is written as 
\begin{align}
&\IM(\H^{(0)} ; \H^{(l)}) \rightarrow \notag\\&\frac{1}{nm}\sum_{i=1}^{n}\sum_{j=1}^{m} D_{K L}\left(\operatorname{Bernoulli}\left(h_{ij}^{(l)}\right) \| \operatorname{Bernoulli}(0.5)\right)
\end{align}

		\begin{algorithm}[t]
			\caption{Framework of the implementation}
			\KwData{dataset $\DM:=\{\H^{(0)},\X\}$; $\y$; hyperparameters $\alpha$, $\beta$, $\epsilon$.}
			\KwResult{$\hat{\y}^{(L)}$.}
			Initializing weights $\phi_1(i)$, $\phi_2(i)$, for $i \in [1,n\_head]$,$\Theta_\text{hid}$,$\Theta_\text{out}$; $ \Z^{(0)}_{\VM} \leftarrow \X$\;
			\For{layers $l=1,\dots, L$}{$\tilde{\H}^{(l)}\leftarrow\mathbf{0}$\;\For{$v \in [1,n]$ and $e \in [1,m]$}{$\A^{l}(v,e)\leftarrow A_{\Z_v^{(l-1)},\H^{(l-1)}}(v,e)$ using Eq. (\ref{eq:attention})\; 
					\If{$\A^{l}(v,e) \leq \epsilon$ }{$\tilde{\H}^{(l)}(v,e)\leftarrow \A^{l}(v,e)$\;}}
				${\H}^{(l)}\leftarrow \{\H^{(0)},\tilde{\H}^{(l)}\}$ using Eq. (\ref{eq:combine})\;
				$	\Z_{\VM}^{(l)} \leftarrow \operatorname{HGNN}(\X,\H^{(l)})$ using Eq. (\ref{eq:hid})\;
				$	\hat{\y}^{(l)} \leftarrow \operatorname{HGNN}(\Z_{\VM}^{(l)},\H^{(l)})$  using Eq. (\ref{eq:out})\;
				$	\LM^{(l)} \leftarrow \operatorname{LOSS}(\hat{\y}^{(l)},\y)$ using Eq. (\ref{eq:loss})\;}
			$\LM\leftarrow\frac{1}{L}\sum_{l=1}^{L}\LM^{(l)}$;
			Back-propagate $\LM$ to update model weights.
		\end{algorithm}

\begin{figure}[] 
	\centering 
	\includegraphics[width=0.9\linewidth]{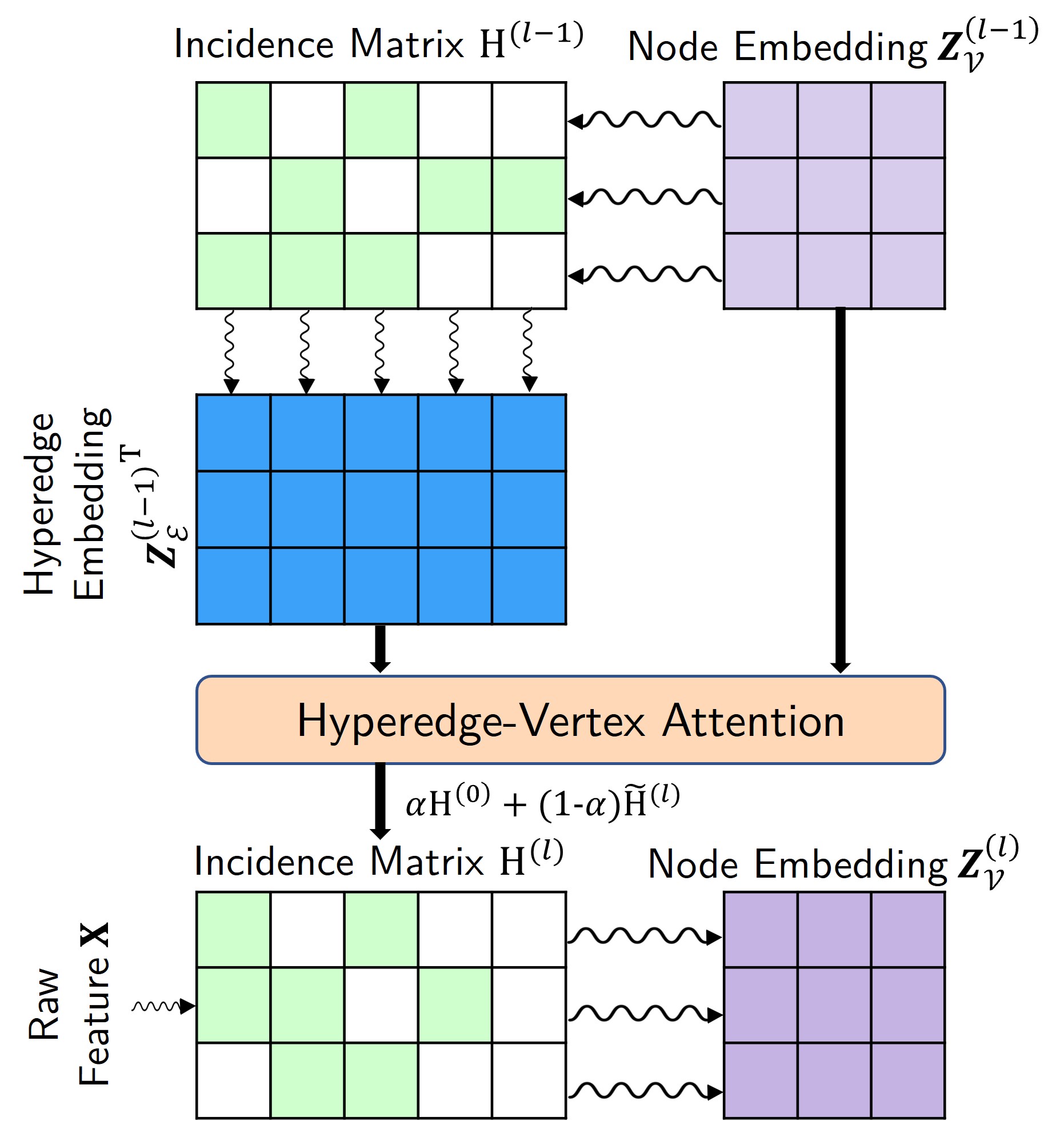} 
	\caption{The pipeline of updating hypergraph structure and node embedding in one layer with 3 vertices and 5 hyperedges.} 
	\label{Fig.attention} 
\end{figure}
\section{An Implementation of the General Paradigm} \label{sc:implementation}

In this section, we give an implementation of the whole network (Algorithm 1). We introduce a hyperedge-vertex attention mechanism to realize the HSL module and the spatial convolution operation to realize the HGNN module.

\subsection{Implementation of Modules} 
The implementation of the HSL module is based on hyperedge-vertex attention. The information of a hyperedge is the aggregation of information of the vertices it contains according to the hyperedge-dependent vertex weight. The attention mechanism is to make a hyperedge attend over all nodes to measure the similarity of information between nodes and hyperedges. Thus, the multi-head attention score between vertex $v$ and hyperedge $e$ is defined as
\begin{align}
   & \z_e  = \sum_{u \in e}\frac{\H(u,e)\z_u}{d(e)} \\
&	A(v,e)  = \frac{1}{K}\sum_{i=1}^{K} \operatorname{sim}\left(\z_v\odot\phi(i), \z_e\odot\phi(i)\right) \label{eq:attention}
\end{align}
where $K$ is the number of attention heads and $\phi(i)$ is the $i$-th attention head weights. 

Empirically, a dense hypergraph structure will contain too much noise and destroy the performance. Therefore, we use a threshold $\epsilon$ to mask the attention scores to reserve the intersections between highly similar hyperedge-vertex pairs. And then, these reserved attentions formulate a new hypergraph $\tilde{\H}$, which is further combined with the initial hypergraph $\H^{(0)}$. The hypergraph structure is updated in the HSL module according to
\begin{align}
	\tilde{\H}^{(l)}(v,e) &= \operatorname{Mask}(	A_{\Z_v^{(l-1)},\H^{(l-1)}}(v,e));\notag\\ {\H}^{(l)} &= \alpha\H^{(0)}+(1-\alpha)\tilde{\H}^{(l)}, \label{eq:combine}
\end{align}
where $\alpha$ is a trade-off hyperparameter to balance the weights of the initial and new structure. 

The HGNN and output modules are each implemented by a spatial hypergraph convolutional layer as in \cite{gao2022hgnn}. The message passing formulation of the $l$-th $(l=1,2,\dots,L)$ iteration is
\begin{align}
\Z_{\VM}^{(l)}& = g_{\mathbf{\Theta}_{1}}(\H^{(l)}, \X) \notag\\
&= \D_v(\H^{(l)})^{-1}\H^{(l)}\D_e(\H^{(l)})^{-1}{\H^{(l)}}^\top\X\mathbf{\Theta}_{1}, \\
	\hat{\y}^{(l)} &= g_{\mathbf{\Theta}_{2}}(\H^{(l)}, \Z_{\VM}^{(l)}) \notag\\
	&=\D_v(\H^{(l)})^{-1}\H^{(l)}\D_e(\H^{(l)})^{-1}{\H^{(l)}}^\top\Z_{\VM}^{(l)}\mathbf{\Theta}_{2}.
\end{align}
In this way, we obtain the hypergraph structure and node embedding of the current layer from those of the previous layer. The pipeline of the proposed implementation is illustrated in Fig. \ref{Fig.attention}.

\subsection{Loss Function}  Plugging the estimations of both mutual information terms in Eq. (\ref{eq:obj}), we obtain the loss function $\LM^{(l)}$:
\begin{align}
\label{eq:loss}
	&\LM_{\text{HIB}}^{(l)} = \sum_{v \in V}  CE \left(\hat{\y}_v^{(l)} ; Y_{v}\right)  \notag \\&+  \beta  
	 \frac{1}{nm}\sum_{i=1}^{n}\sum_{j=1}^{m} D_{K L}\left(\operatorname{Bernoulli}\left(\H_{ij}^{(l)}\right) \| \operatorname{Bernoulli}(0.5)\right)
\end{align}

The final loss is obtained by adding up the losses for $L$ iterations $\LM = \sum_{l=1}^{L}\LM_{\text{HIB}}^{(l)}$. The gradient back-propagation algorithm is used to update the parameters of the entire network. 

\subsection{Convergence Analysis}
Here we first prove that the mutual information between $\H^{(0)}$ and $\H^{(l)}$ is descending from layer to layer in the feed-forward process in each epoch.

\textbf{Proposition 3.} In our implementation, $\IM(\H^{(0)},\H^{(l)})<\IM(\H^{(0)},\H^{(l+1)})$.

\textbf{Proof.} We bound $\IM(\H^{(0)},\H^{(l+1)})$ by

\begin{align}
    &\IM(\H^{(0)},\H^{(l+1)}) \notag\\ &\overset{1)}{=} H(\H^{(0)})+H(\H^{(l+1)})-H(\H^{(0)},\H^{(l+1)}) \notag\\
    &\overset{2)}{=}H(\H^{(0)})+\sum_{\H^{(0)},\H^{(l+1)}}\mathbb{P}(\H^{(l+1)})\log \frac{\mathbb{P}(\H^{(l+1)})}{\mathbb{P}(\H^{(0)})} \notag\\
    &\overset{3)}{\leq} H(\H^{(0)}) + (1-\alpha)\sum_{\H^{(0)},\tilde{\H}^{(l+1)}}\mathbb{P}(\tilde{\H}^{(l+1)})\log \frac{\mathbb{P}(\tilde{\H}^{(l+1)})}{\mathbb{P}(\H^{(0)})} \notag\\
    &\overset{4)}{<} H(\H^{(0)}) + \sum_{\H^{(0)},\tilde{\H}^{(l+1)}}\mathbb{P}(\tilde{\H}^{(l+1)})\log \frac{\mathbb{P}(\tilde{\H}^{(l+1)})}{\mathbb{P}(\H^{(0)})} \notag\\
    &\overset{5)}{=}\IM(\H^{(0)},\tilde{\H}^{(l+1)}) \label{eq:proof3.1}
\end{align}
\begin{table*}[]
	\centering
	\caption{Details for all datasets used in the experiments.} \label{tb:detail}
	\begin{tabular}{lccccccc}
		\hline
	Datasets	& \#Nodes & \#Hyperedges & \#Features & \#classes & \#Training & \#Validation & \#Testing \\ \hline
		Cora      & 2708    & 2708         & 1433       & 7        & 140 & 500        & 1000      \\
		Citeseer  & 3327    & 3327         & 3703       & 6        & 120  & 500        & 1000    \\
		Yummly10k & 10149   & 1269         & 512        & 7        & 1400   & 210        & 8539   \\
		DBLP-coauthor & 5063   & 3607         & 768        & 3        & 300   & 150        & 4613   \\ \hline
	\end{tabular}
\end{table*}
where 1), 2), 5) is the definition of mutual information, 3) uses Eq. (\ref{eq:combine}) and Jensen's inequality of convex function and 4) uses $0<\alpha<1$.

Next, the conditional distribution of $\tilde{\H}^{(l+1)}$ depends only on $\H^{(l)}$ while independent of $\H^{(0)}$ in our implementation. Thus, we have the Markov chain in the order $\H^{(0)} \to \H^{(l)} \to \H^{(l+1)}$. Based on Eq. (\ref{eq:proof3.1}) and the data-processing inequality, the following inequality holds:
\begin{equation}
	\IM(\H^{(0)},\H^{(l+1)}) < \IM(\H^{(0)} ,  \tilde{\H}^{(l+1)}) \leq \IM(\H^{(0)} , \H^{(l)})
	\label{eq:proof3.2}
\end{equation}

\begin{figure}[] 
	\centering 
	\includegraphics[width=\linewidth]{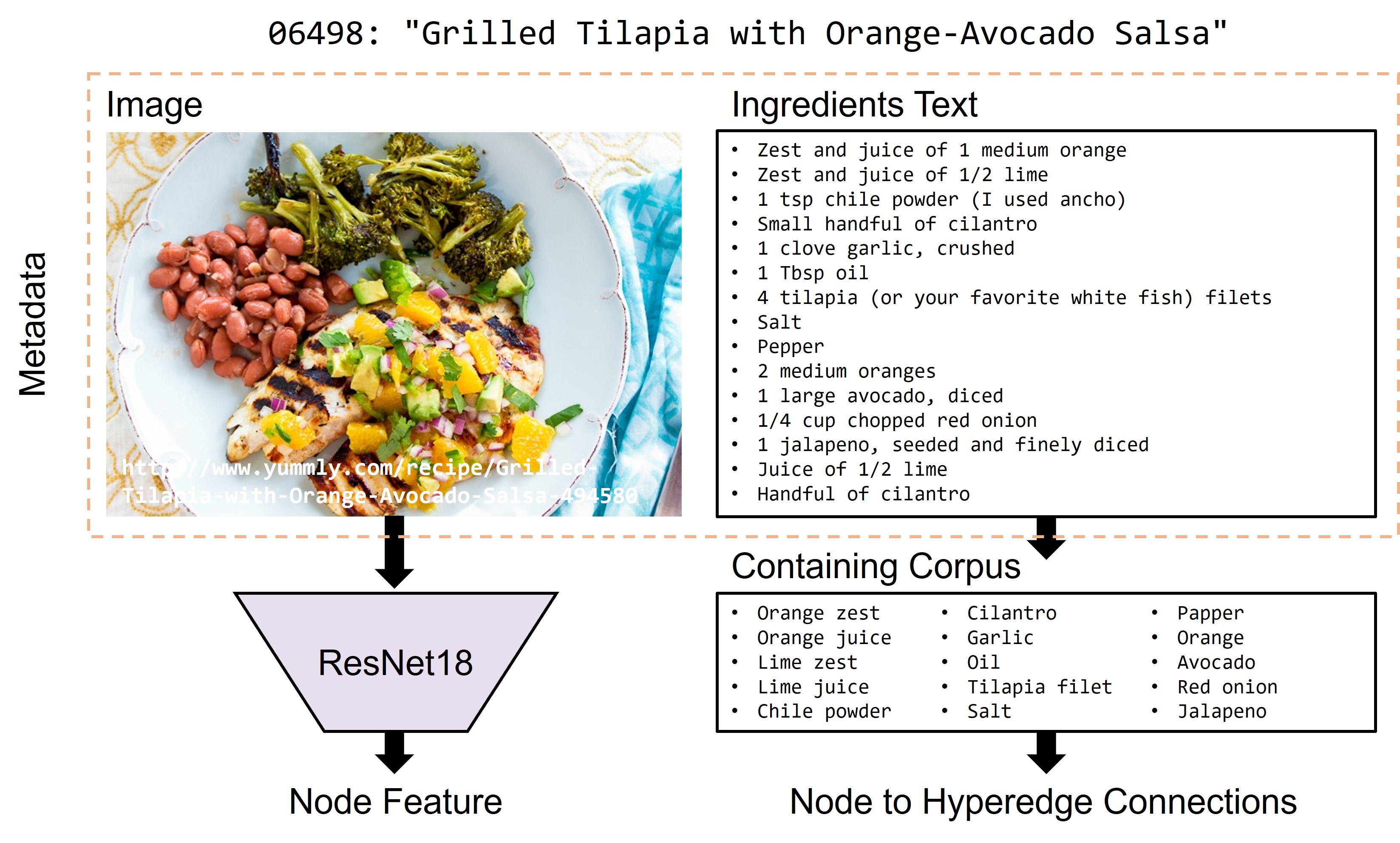} 
	\caption{Preprocessing pipeline of Yummly10k dataset.} 
	\label{Fig.pipeline} 
\end{figure}

 Obviously, $\IM(\H^{(0)} , \H^{(l)})$ has finite upper bound and lower bound, which are the information entropy of $\H^{(0)}$ and zero, respectively. Therefore, the mutual information $\IM(\H^{(0)} ; \H^{(l)})$ will converge to a finite value in each epoch (inner loop). Furthermore, since the loss function empirically tends to converge during the training process, its second term, the estimation of $\IM(\H^{(0)} ; \H^{(l)})$, is inclined to be stable in the outside loop. Summarizing the above two points, we can get to the conclusion that the mutual information between the new structure and the initial structure will generally converge in the training phase.



\section{Experiments} \label{sc:experiement}

The aim of our experiments is to show the effectiveness and robustness of our method. We will discuss the following four issues in our experiments: (1) the classification accuracy of DeepHGSL compared with state-of-the-art methods; (2) the robustness of DeepHGSL to defend against noisy structure; (3) the contribution of HIB to such robustness; (4) 
the sensitivity of hyperparameters $\alpha$, $\beta$, and $\epsilon $, respectively.

\subsection{Datasets} 


We note that there are few benchmark datasets specifically targeting at node classification task on a hypergraph. Thus, we make a cooking dataset, named Yummly10k, in which nodes are recipes and hyperedges are ingredients. The recipes containing the same ingredient are connected by one hyperedge. We get the raw data of Yummly-28K dataset from the project website \footnote{http://123.57.42.89/FoodComputing\_\_Dataset.html} and conduct a preprocessing pipeline including data selection, feature extraction, and hypergraph construction as shown in Fig. \ref{Fig.pipeline}. we select the ``cuisine'' as the target of classification task. In Yummly-28K, each recipe may contain several cuisine labels, and the domain of origin cuisines intersects with each other.  Similar relations are found in [American, Southern \& Soul Food, Cajun \& Creole], [Asian, Thai, Chinese], [Greek, Mediterranean, Moroccan, Italian], etc. Therefore, we remove those cuisines that contain other cuisines and contain very few recipes, and only reserve these typical cuisines with clear domains including Barbecue, Chinese, French, Indian, Kid-Friendly, Mexican, and Thai. Finally, the Yummly10k with balanced and clear labels is generated. For each cuisine, we randomly select $200$ and $30$ samples for training and validation, and the rest is used for testing.  Features are extracted from the images of recipes by a ResNet18 network \cite{he2016deep} pre-trained on the ImageNet dataset \cite{imagenet_cvpr09} and finetuned on our dataset.

\begin{table}[]
\centering
\caption{Hyperparameter settings for DeepHGSL on all datasets  used in the experiments.} \label{tb:setting}
 \setlength{\tabcolsep}{5mm}{
 \begin{tabular}{lcccc}
\toprule
Datasets  & $\alpha$ & $\beta$ & $\epsilon$ & $L$  \\ \midrule
Cora      &     0.7     &    0.01    &      0      &  5       \\
Citeseer  &     0.8     &    0.01     &      0.1    &  5      \\
Yummly10k &     0.5     &    0.1     &      0.1      &  10       \\
DBLP-coauthor &     0.5     &   1     &      0.1      &  7       \\ \bottomrule
\end{tabular} }
\end{table}

Since the ingredient text in Yummly-28K is unformatted natural language, we first perform simple natural language processing on the raw text and extract the corpus of ingredients. The automatically extracted corpus exist semantic repetition due to different languages, synonyms, spelling errors, inconsistent word order, etc. Therefore, we manually screened the corpus and finally produced a corpus including 1463 ingredient names. 
In order to build a hypergraph, we use each kind of ingredient as a hyperedge and each recipe as a node. For each recipe, we analyze the text of ingredients and match it with the above corpus. If a corpus is matched, then the node belongs to the corresponding hyperedge.
 This analysis covered 88.6\% of ingredients (117,196/132,238). After the hypergraph is established, we remove the top 100 hyperedges with the largest degree and the hyperedges with a degree of 0 to obtain the final hypergraph.


\begin{table*}[] \label{tb:cora}
\setlength\tabcolsep{3pt} 

    \centering
	\caption{Average classification accuracy on Cora dataset (in percent).} \label{tb:cora}
	\begin{tabular}{|p{0.35cm}<{\centering}|p{1.95cm}|p{1.38cm}<{\centering}|p{1.38cm}<{\centering}p{1.38cm}<{\centering}p{1.38cm}<{\centering}|p{1.38cm}<{\centering}p{1.38cm}<{\centering}p{1.38cm}<{\centering}|}
		\hline
		\multirow{2}{*}{}          & \multicolumn{1}{c|}{\multirow{2}{*}{Model}} & Clean & \multicolumn{3}{c|}{Delete edges} & \multicolumn{3}{c|}{Add edges} \\ \cline{3-9} 
		& \multicolumn{1}{c|}{}                       &       & 25\%    & 50\%    & 75\%    & 25\%   & 50\%   & 75\%    \\ \hline
		\multirow{8}{*}{\rotatebox{90}{Cora}} & GCN\cite{gcn}                                         &    $80.6_{\pm1.26}$   &    $78.0_{\pm1.33}$   &  $73.5_{\pm1.63}$       &    $66.7_{\pm1.55}$     &   $76.9_{\pm1.27}$     &  $72.9_{\pm1.79}$      &    $70.6_{\pm1.90}$     \\
		& GAT\cite{gat}                                       &    $80.5_{\pm1.34}$   &   $79.0_{\pm1.11}$      &    $75.0_{\pm2.26}$     &   $68.4_{\pm1.53}$      &   $75.4_{\pm2.14}$     &  $70.9_{\pm3.06}$      &      $67.3_{\pm2.50}$    \\
		& IDGL\cite{chen2020iterative}                                       &   $82.5_{\pm0.42}$    &    $80.2_{\pm0.80}$     &    $77.4_{\pm1.17}$     &  $71.2_{\pm1.04}$      &  $77.2_{\pm0.76}$      &   $73.8_{\pm1.38}$     &    $72.0_{\pm1.53}$      \\ 
		& GIB\cite{wu2020graph}                                     &   $82.0_{\pm0.26}$    &   $79.8_{\pm1.25}$      &  $76.1_{\pm1.42}$       & $70.8_{\pm1.10}$   &  $75.5_{\pm1.23}$      &   $70.0_{\pm1.62}$     &    $63.5_{\pm1.40}$      \\ \cline{2-9} 
		& HyperGCN\cite{yadati2019hypergcn}                                  &   $75.5_{\pm1.54}$    &    $71.4_{\pm2.33}$     & $63.0_{\pm1.95}$        &  $51.2_{\pm3.09}$    &   $61.8_{\pm2.63}$     &  $53.9_{\pm1.59}$      &  $47.8_{\pm2.53}$           \\
		& HCHA\cite{bai2021hypergraph}                                    &   $82.2_{\pm1.99}$    &    $74.6_{\pm1.52}$     &    $67.9_{\pm1.74}$     &   $59.2_{\pm3.69}$   &   $66.2_{\pm1.91}$     &  $59.6_{\pm2.31}$      &    $54.9_{\pm3.43}$         \\ 
		& HGNN$^{+}$\cite{gao2022hgnn}                                     &       $80.9_{\pm0.57}$ &   $77.8_{\pm1.43}$      &     $73.4_{\pm1.63}$    &  $67.5_{\pm1.02}$       &   $73.6_{\pm1.00}$     &   $68.4_{\pm1.50}$     &   $64.3_{\pm2.19}$      \\ \cline{2-9} 
		& DeepHGSL                                         &       \textbf{$\mathbf{83.5_{\pm0.80}}$}&    \textbf{$\mathbf{81.3_{\pm0.95}}$}     &    \textbf{$\mathbf{78.6_{\pm0.63}}$}     &  \textbf{$\mathbf{73.5_{\pm1.37}}$}       &       \textbf{$\mathbf{78.0_{\pm0.68}}$} &     \textbf{$\mathbf{75.2_{\pm0.79}}$}   &  \textbf{$\mathbf{73.6_{\pm1.05}}$}         \\
		\hline
	\end{tabular}

\end{table*}

\begin{table*}[] 
\setlength\tabcolsep{3pt} 
    \centering

	\caption{Average classification accuracy on Citeseer dataset (in percent).} \label{tb:citeseer}
	\begin{tabular}{|p{0.35cm}<{\centering}|p{1.95cm}|p{1.38cm}<{\centering}|p{1.38cm}<{\centering}p{1.38cm}<{\centering}p{1.38cm}<{\centering}|p{1.38cm}<{\centering}p{1.38cm}<{\centering}p{1.38cm}<{\centering}|}
		\hline
		\multirow{2}{*}{}          & \multicolumn{1}{c|}{\multirow{2}{*}{Model}} & Clean & \multicolumn{3}{c|}{Delete edges} & \multicolumn{3}{c|}{Add edges} \\ \cline{3-9} 
		& \multicolumn{1}{c|}{}                       &       & 25\%    & 50\%    & 75\%    & 25\%   & 50\%   & 75\%    \\ \hline
		\multirow{8}{*}{\rotatebox{90}{Citeseer}} & GCN\cite{gcn}                                       &    $70.6_{\pm0.76}$   &    $68.8_{\pm0.95}$   &  $65.7_{\pm1.84}$       &    $61.0_{\pm2.52}$     &   $65.8_{\pm1.48}$     &  $62.5_{\pm1.43}$      &    $60.0_{\pm1.46}$     \\
		& GAT\cite{gat}                                    &   $71.8_{\pm0.75}$    &    $69.8_{\pm1.45}$     &     $65.4_{\pm4.35}$    &   $63.9_{\pm2.39}$      &   $66.0_{\pm2.63}$     &    $62.2_{\pm1.22}$    &  $60.4_{\pm1.82}$        \\
		& IDGL\cite{chen2020iterative}                                    &   $72.2_{\pm0.98}$    &    $71.0_{\pm1.35}$     &    $69.1_{\pm1.71}$     &   $67.9_{\pm1.76}$     &    $69.2_{\pm1.43}$    &   \textbf{$\mathbf{67.4_{\pm1.57}}$}     &    $65.8_{\pm1.73}$      \\
			& GIB\cite{wu2020graph}                                    &    $67.7_{\pm0.85}$   &   $65.8_{\pm1.92}$      & $62.9_{\pm1.46}$        &   $53.7_{\pm9.01}$     &  $63.5_{\pm0.76}$      &    $58.4_{\pm0.29}$    &    $54.7_{\pm0.42}$      \\ \cline{2-9} 
		& HyperGCN\cite{yadati2019hypergcn}                                   &   $59.1_{\pm2.07}$    &     $57.0_{\pm2.00}$    &     $54.4_{\pm1.65}$    &   $49.4_{\pm2.87}$   &   $48.9_{\pm2.24}$     &   $43.2_{\pm1.90}$     &    $38.8_{\pm2.13}$         \\
		& HCHA\cite{bai2021hypergraph}                                   &    $70.8_{\pm1.19}$    &   $61.7_{\pm1.50}$       &     $58.5_{\pm2.19}$     &    $52.3_{\pm3.02}$   &   $51.5_{\pm2.24}$      &   $44.9_{\pm2.90}$      &    $39.7_{\pm1.71}$          \\ 
		& HGNN$^{+}$\cite{gao2022hgnn}                                    &       $70.4_{\pm0.50}$ &   $68.6_{\pm0.84}$      &     $66.2_{\pm1.70}$    &  $61.7_{\pm1.41}$       &   $63.9_{\pm1.42}$     &   $59.4_{\pm1.62}$     &   $55.7_{\pm1.98}$      \\ \cline{2-9} 
		& DeepHGSL                                         &       \textbf{$\mathbf{72.8_{\pm0.68}}$}&    \textbf{$\mathbf{71.6_{\pm0.51}}$}     &    \textbf{$\mathbf{70.4_{\pm0.98}}$ }    &  \textbf{$\mathbf{68.6_{\pm1.28}}$}       &    \textbf{$\mathbf{69.4_{\pm1.22}}$}     &    $67.3_{\pm1.37}$    &    \textbf{$\mathbf{66.1_{\pm1.38}}$ }      \\
		\hline
	\end{tabular}

\end{table*}
  
We also conduct experiments on the co-author DBLP dataset (DBLP-HG) \cite{hu2021adaptive}, in which the co-authorship is used to build the hyperedges. The DBLP-HG dataset includes 5063 nodes and 3607 hyperedges. The papers are classified into three categories. For each kind of paper, $100$ and $50$ samples per class are randomly selected for training and validation. To demonstrate the extension of our method on graph-structured data, We conduct experiments on the commonly used open benchmark graph datasets: Cora and Citeseer. In these citation networks, each node represents a document and each edge represents citation links. We follow the standard train-validation-test split as GCN \cite{gcn}.  The details for all four datasets are summarized in Table \ref{tb:detail}.

\subsection{Experimental Set-up} 

To study the robustness of different methods under structure perturbations, we conduct experiments with the following four experimental settings.

\begin{itemize}
    \item[1] Clean. The original graph/hypergraph structure is used.
    \item[2] Delete edges/hyperedges. Randomly delete 25\%, 50\%, and 75\% of the edges/hyperedges in the original structure. 
    \item[3] Add edges on Cora and Citeseer datasets. If the original graph structure has $n$ nodes and $m$ edges, we randomly select $25\%m$, $50\%m$, and $75\%m$ pairs from the $n^2-m$ unconnected pairs of nodes and add edges.
    \item[4] Add hyperedges on Yummly10k and DBLP-coauthor datasets. The number of hyperedges in the original hypergraph structure is $m$. For each time, we select one sample from each class with equal probability and the set of selected nodes forms a hyperedge. This procedure is repeated multiple times until $25\%m$, $50\%m$, and $75\%m$ hyperedges are added. 
\end{itemize}

 All models are trained for a maximum of 10,000 epochs using Adam optimizer with a learning rate of 0.01. The training will be early stopped if the validation accuracy does not increase for 500 consecutive epochs. we repeat the experiments with 10 random initializations and structure perturbations and report average performance. Hyperparameters are tuned on the validation set.
 Table \ref{tb:setting} shows the hyperparameter settings of our method on Cora, Citeseer and Yummly10k datasets. $L$ is the number of layers in the general paradigm.  In all our experiments, the dropout ratio and the size of the hidden layer are set to 0.5 and 16, respectively.  The number of attention heads in the HSL module is set to 6 in all experiments.

\begin{table*}[h] 
\setlength\tabcolsep{3pt} 
    \centering
	\caption{Average classification accuracy on Yummly10k dataset (in percent).} \label{tb:yummly10k}
	\begin{tabular}{|p{0.35cm}<{\centering}|p{1.95cm}|p{1.38cm}<{\centering}|p{1.38cm}<{\centering}p{1.38cm}<{\centering}p{1.38cm}<{\centering}|p{1.38cm}<{\centering}p{1.38cm}<{\centering}p{1.38cm}<{\centering}|}
		\hline
		\multirow{2}{*}{}          & \multicolumn{1}{c|}{\multirow{2}{*}{Model}} & Clean & \multicolumn{3}{c|}{Delete hyperedges} & \multicolumn{3}{c|}{Add hyperedges} \\ \cline{3-9} 
		& \multicolumn{1}{c|}{}                       &       & 25\%    & 50\%    & 75\%    & 25\%   & 50\%   & 75\%    \\ \hline
		\multirow{7}{*}{\rotatebox{90}{Yummly10k}} 
		& MLP                                         &  $37.1_{\pm2.27}$     &    ---     &    ---     &   ---     &    ---    &   ---     &     ---     \\
		& GCN\cite{gcn}                                 &    $74.2_{\pm3.31}$   &    $65.6_{\pm4.54}$   &  $56.3_{\pm2.38}$       &    $36.6_{\pm5.24}$     &   $71.8_{\pm1.62}$     &  $73.1_{\pm1.16}$      &    $72.2_{\pm1.40}$     \\
		& GAT\cite{gat}                                &   $69.7_{\pm0.31}$    &    $63.0_{\pm1.31}$     &   $53.2_{\pm1.50}$      &   $42.4_{\pm1.30}$      &   $68.9_{\pm0.72}$   &   $68.2_{\pm0.88}$     &    $67.9_{\pm0.49}$      \\
		 \cline{2-9} 
		& HyperGCN\cite{yadati2019hypergcn}                                   &   $46.3_{\pm1.68}$    &      $45.1_{\pm1.99}$    &     $44.0_{\pm2.62}$     &    $41.4_{\pm2.22}$   &      $45.8_{\pm1.42}$   &     $43.8_{\pm2.17}$    &       $41.3_{\pm3.65}$       \\
		& HCHA\cite{bai2021hypergraph}                                   &   $43.9_{\pm1.91}$    &   $42.7_{\pm2.47}$      &   $41.6_{\pm1.64}$      &  $41.0_{\pm2.61}$    &   $44.4_{\pm2.77}$     &    $44.3_{\pm2.60}$    &     $43.9_{\pm3.52}$        \\ 
		& HGNN$^{+}$\cite{gao2022hgnn}                                    &       $75.4_{\pm2.69}$ &   $66.5_{\pm2.55}$      &     $56.6_{\pm2.47}$    &  $37.1_{\pm3.59}$       &   $74.3_{\pm1.12}$     &   $74.1_{\pm1.52}$     &   $72.4_{\pm1.29}$      \\ \cline{2-9} 
		& DeepHGSL&       \textbf{$\mathbf{79.3_{\pm0.85}}$}&    \textbf{$\mathbf{74.1_{\pm2.28}}$}     &    \textbf{$\mathbf{68.3_{\pm4.72}}$}     &  \textbf{$\mathbf{53.7_{\pm3.14}}$}       &       \textbf{$\mathbf{79.1_{\pm0.85}}$} &     \textbf{$\mathbf{74.5_{\pm1.24}}$}   &  \textbf{$\mathbf{74.0_{\pm1.15}}$}         \\
		\hline
	\end{tabular}

\end{table*}

\begin{table*}[h] 
\setlength\tabcolsep{3pt} 
    \centering
	\caption{Average classification accuracy on DBLP-coauthor dataset (in percent).} \label{tb:dblp}
	\begin{tabular}{|p{0.35cm}<{\centering}|p{1.95cm}|p{1.38cm}<{\centering}|p{1.38cm}<{\centering}p{1.38cm}<{\centering}p{1.38cm}<{\centering}|p{1.38cm}<{\centering}p{1.38cm}<{\centering}p{1.38cm}<{\centering}|}
		\hline
		\multirow{2}{*}{}          & \multicolumn{1}{c|}{\multirow{2}{*}{Model}} & Clean & \multicolumn{3}{c|}{Delete hyperedges} & \multicolumn{3}{c|}{Add hyperedges} \\ \cline{3-9} 
		& \multicolumn{1}{c|}{}                       &       & 25\%    & 50\%    & 75\%    & 25\%   & 50\%   & 75\%    \\ \hline
		\multirow{7}{*}{\rotatebox{90}{DBLP-coauthor}} 
		& MLP                                         &  $71.8_{\pm1.24}$     &    ---     &    ---     &   ---     &    ---    &   ---     &     ---     \\
		& GCN\cite{gcn}                                 &    $73.5_{\pm0.23}$   &    $73.5_{\pm1.63}$   &  $72.2_{\pm1.03}$       &    $71.8_{\pm1.44}$     &   $71.6_{\pm0.42}$     &  $70.9_{\pm0.63}$      &    $69.2_{\pm0.41}$     \\
		& GAT\cite{gat}                                &  $65.7_{\pm0.27}$    &    $64.4_{\pm0.68}$     &   $63.6_{\pm0.77}$      &   $62.7_{\pm1.06}$      &    $65.2_{\pm0.56}$    &   $64.1_{\pm0.27}$    &   $63.3_{\pm0.34}$      \\
		 \cline{2-9} 
		& HyperGCN\cite{yadati2019hypergcn}                                   &   $68.1_{\pm0.95}$    &      $66.8_{\pm0.53}$    &     $69.1_{\pm0.12}$     &    $70.6_{\pm0.46}$   &      $61.9_{\pm0.56}$   &     $55.3_{\pm0.60}$    &       $52.4_{\pm1.64}$       \\
		& HCHA\cite{bai2021hypergraph}                                   &   $73.1_{\pm0.52}$    &   $72.6_{\pm1.58}$      &   $72.6_{\pm0.12}$      &  $71.0_{\pm2.09}$    &   $70.2_{\pm1.60}$     &    $67.7_{\pm1.30}$    &     $63.3_{\pm0.25}$        \\ 
		& HGNN$^{+}$\cite{gao2022hgnn}                                    &       $74.3_{\pm1.12}$ &   $71.7_{\pm1.19}$      &     $70.6_{\pm0.46}$    &  $69.5_{\pm1.55}$       &   $70.5_{\pm0.41}$     &   $68.3_{\pm0.28}$     &   $67.4_{\pm0.42}$      \\ \cline{2-9} 
		& DeepHGSL                                         &       \textbf{$\mathbf{77.2_{\pm0.32}}$}&    \textbf{$\mathbf{76.6_{\pm0.30}}$}     &    \textbf{$\mathbf{76.1_{\pm0.23}}$}     &  \textbf{$\mathbf{75.2_{\pm0.59}}$}       &       \textbf{$\mathbf{75.8_{\pm0.59}}$} &     \textbf{$\mathbf{75.5_{\pm0.29}}$}   &  \textbf{$\mathbf{75.4_{\pm0.29}}$}         \\
		\hline
	\end{tabular}

\end{table*}

 We compare our method, DeepHGSL, with the cutting-edge graph and hypergraph neural networks including GCN \cite{gcn}, HyperGCN \cite{yadati2019hypergcn}, and HGNN$^{+}$ \cite{gao2022hgnn}. Considering that the attention mechanism is applied in our implementation, we also compare DeepHGSL with two approaches, GAT \cite{gat} and HCHA \cite{bai2021hypergraph}, that introduce the attention mechanism into graph and hypergraph neural networks. We further compare against two deep graph structure learning algorithms, \ie, iterative deep graph learning \cite{chen2020iterative} (IDGL) and graph information bottleneck \cite{wu2020graph} (GIB). The clique expansion \cite{agarwal2006higher} is used to transform a hypergraph into a graph to apply the graph learning methods to the hypergraph datasets. For transformations from graphs to hypergraphs, we construct a hypergraph from a citation network by connecting each node and its one-hop neighborhoods by a hyperedge.

\subsection{Experimental Results and Discussions}

The classification accuracy on Cora, Cite, Yummly10k, and DBLP-coauthor datasets are summarized in Tables \ref{tb:cora}, \ref{tb:citeseer}, \ref{tb:yummly10k}, and \ref{tb:dblp}. As demonstrated here, our method outperforms the compared methods by a significant margin in most cases. For graph-structured data, the classification accuracy of DeepHGSL is 3.6\% and 3.12\% relatively higher than that of the GCN method  on clean Cora and Citeseer datasets, respectively. We note that it barely improves the performance by converting the initial graph to hypergraph structure and subsequently using hypergraph-based learning methods. For example, the performance of HGNN$^{+}$ and GCN is roughly close. However, our method can achieve better performance than the graph structure learning methods like IDGL and GIB. This is because the search space of the hypergraph structure is much larger than that of the graph structure with the same number of nodes, and the structure is more fully optimized in our method. 

DeepHGSL can effectively optimize the initial hypergraph structure and boost the performance of the basic network HGNN$^{+}$. For example, DeepHGSL outperforms HGNN$^{+}$ by 3.9\% on both clean Yummly10k and DBLP-coauthor datasets. We note that methods aggregating information based on attention mechanism, such as GAT and HCHA, are sensitive to hypergraph-structured data, and their performance fluctuates greatly due to the fact that the aggregator may fail to attend over really useful connections when there are too many neighborhoods for a node in the hypergraph. As a comparison, methods without learning attention weights, such as GCN and HGNN$^{+}$, are more stable on hypergraph-structured data. However, when too many hyperedges are removed, these methods will also break down because the rest hypergraph structure is disconnected. Our method uses attention mechanism to obtain a new hypergraph structure and 
combines it with the original structure to avoid drastic fluctuations when updating the structure. Another observation is that HyperGCN has poor performance on both hypergraph datasets, 
which  may be explained by the information lost during the conversion from hypergraph structure to graph structure.

 \begin{figure*}[]
	\centering
	\subfigure[Yummly10k]{
			\includegraphics[width=0.45\linewidth]{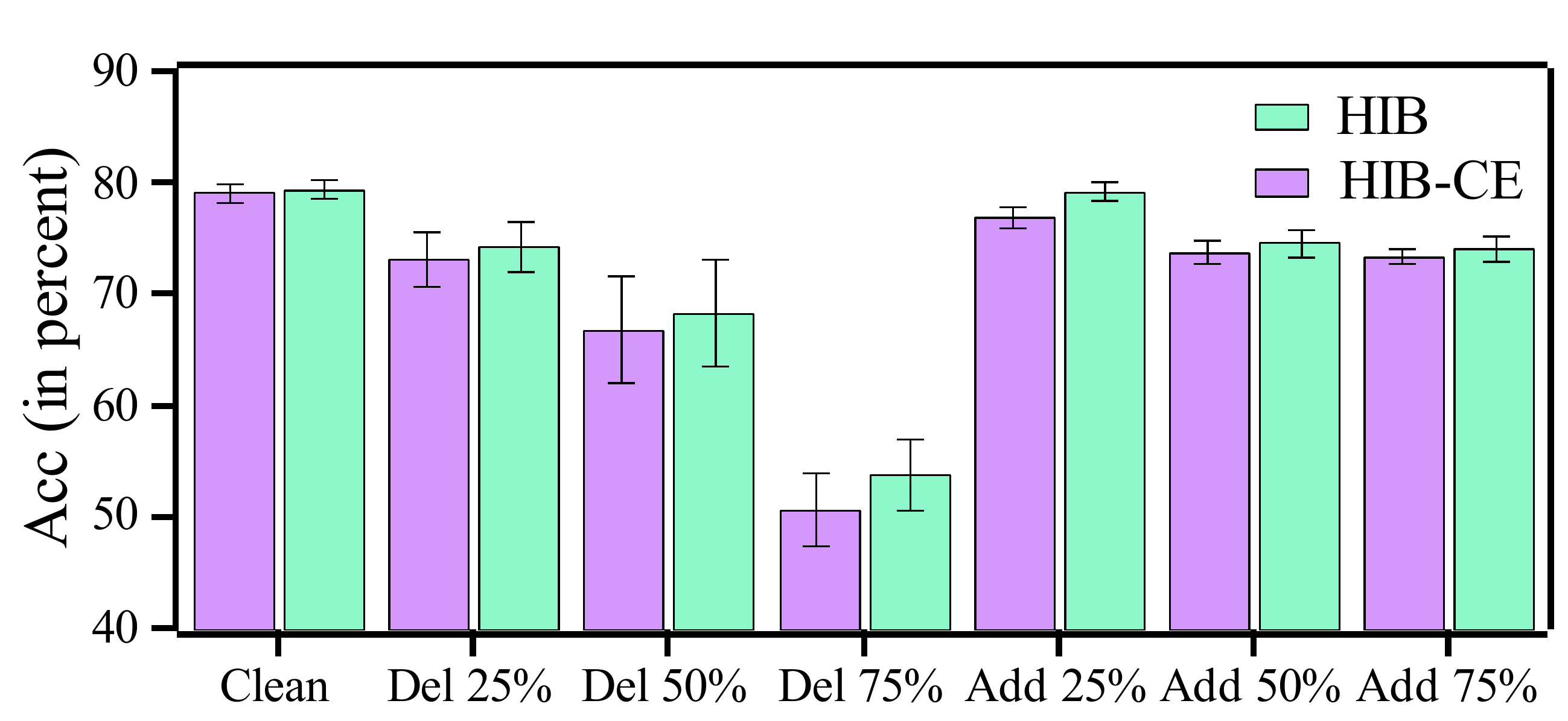}
	}%
	\subfigure[DBLP-coauthor]{
			\includegraphics[width=0.45\linewidth]{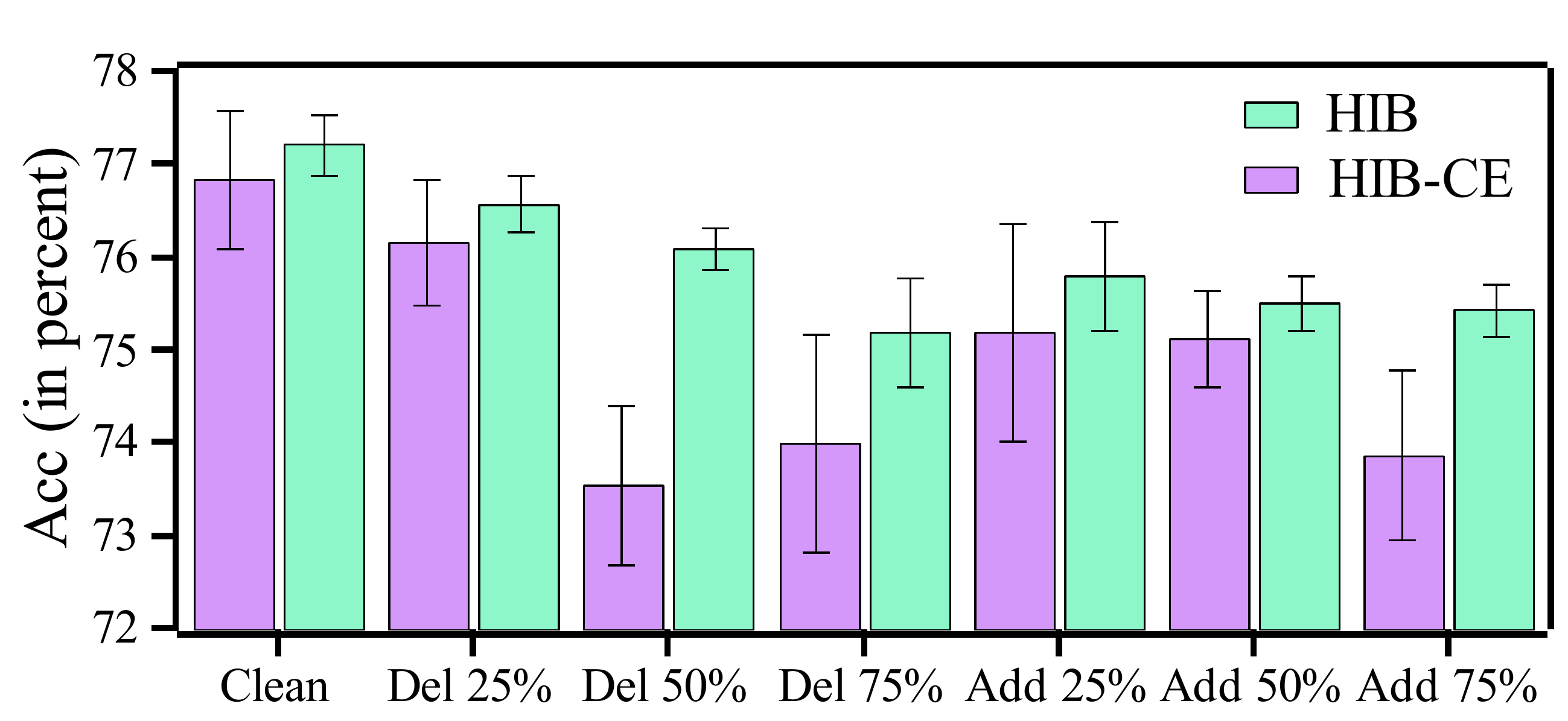}
	}%
	\centering
	\caption{Average classification accuracy for the ablations of DeepHGSL on Yummly10k and DBLP-coauthor datasets (in percent).}\label{fig:abalation}
\end{figure*}

\begin{figure}[]
	\centering
	\includegraphics[width=0.85\linewidth]{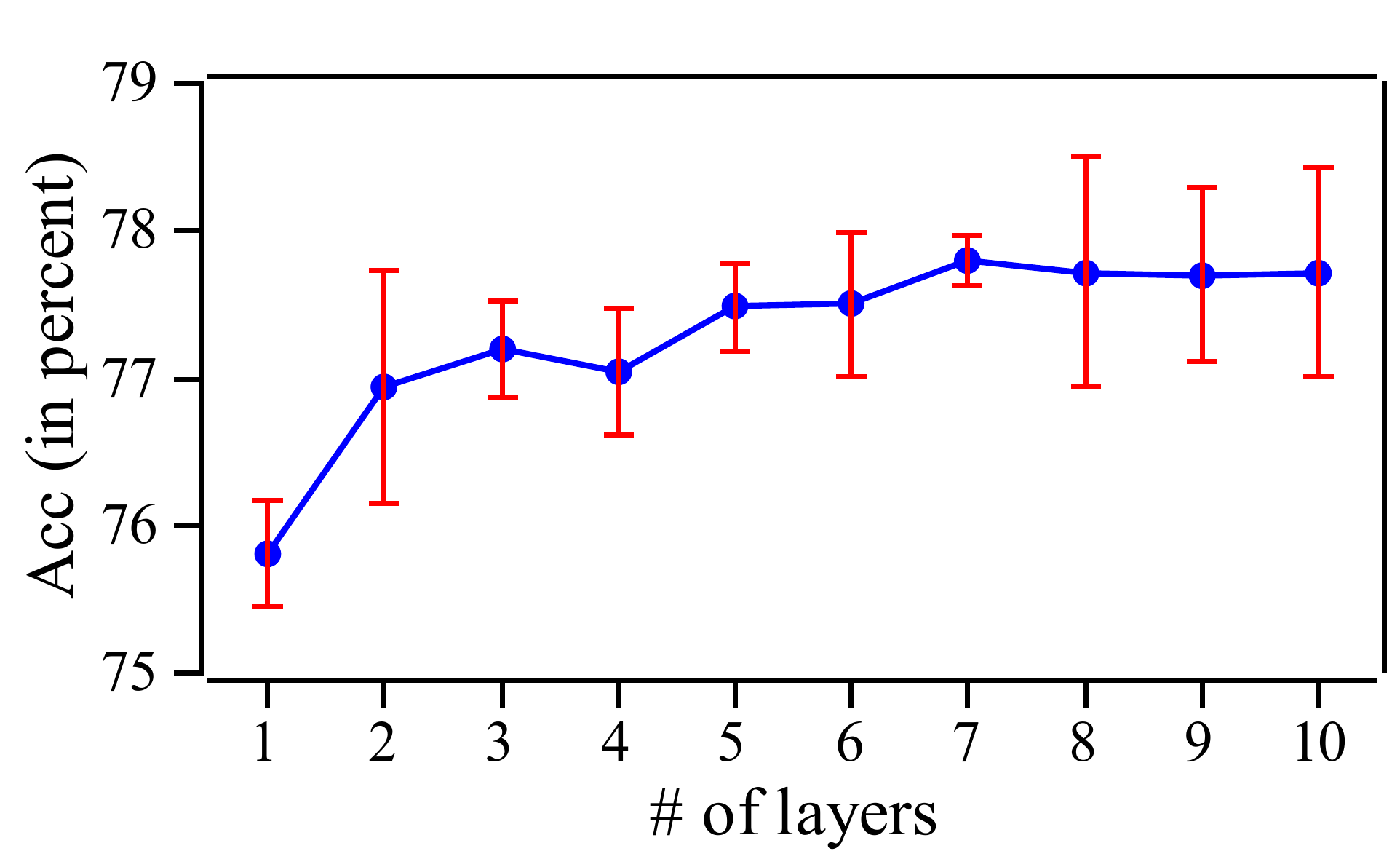}
	\caption{The average classification accuracy of DeepHGSL by varying the number of layers on clean DBLP-coauthor dataset .}\label{fig:abalation-layer}
\end{figure}

\begin{figure*}[htbp]
	\centering
	\subfigure{
		\begin{minipage}[t]{0.3\linewidth}
			\centering
			\includegraphics[width=0.95\linewidth]{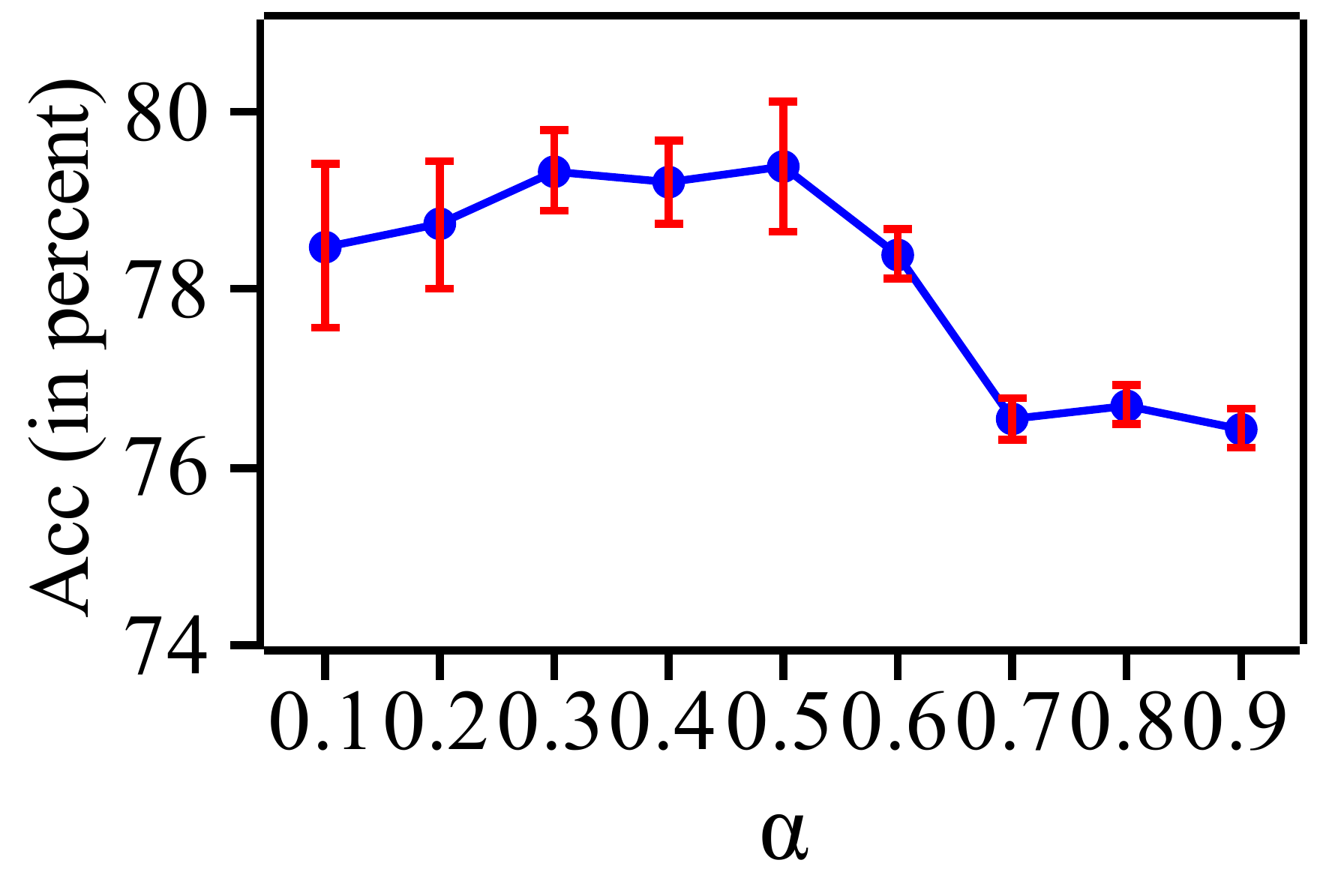}
		\end{minipage}%
	}%
	\subfigure{
		\begin{minipage}[t]{0.3\linewidth}
			\centering
			\includegraphics[width=0.95\linewidth]{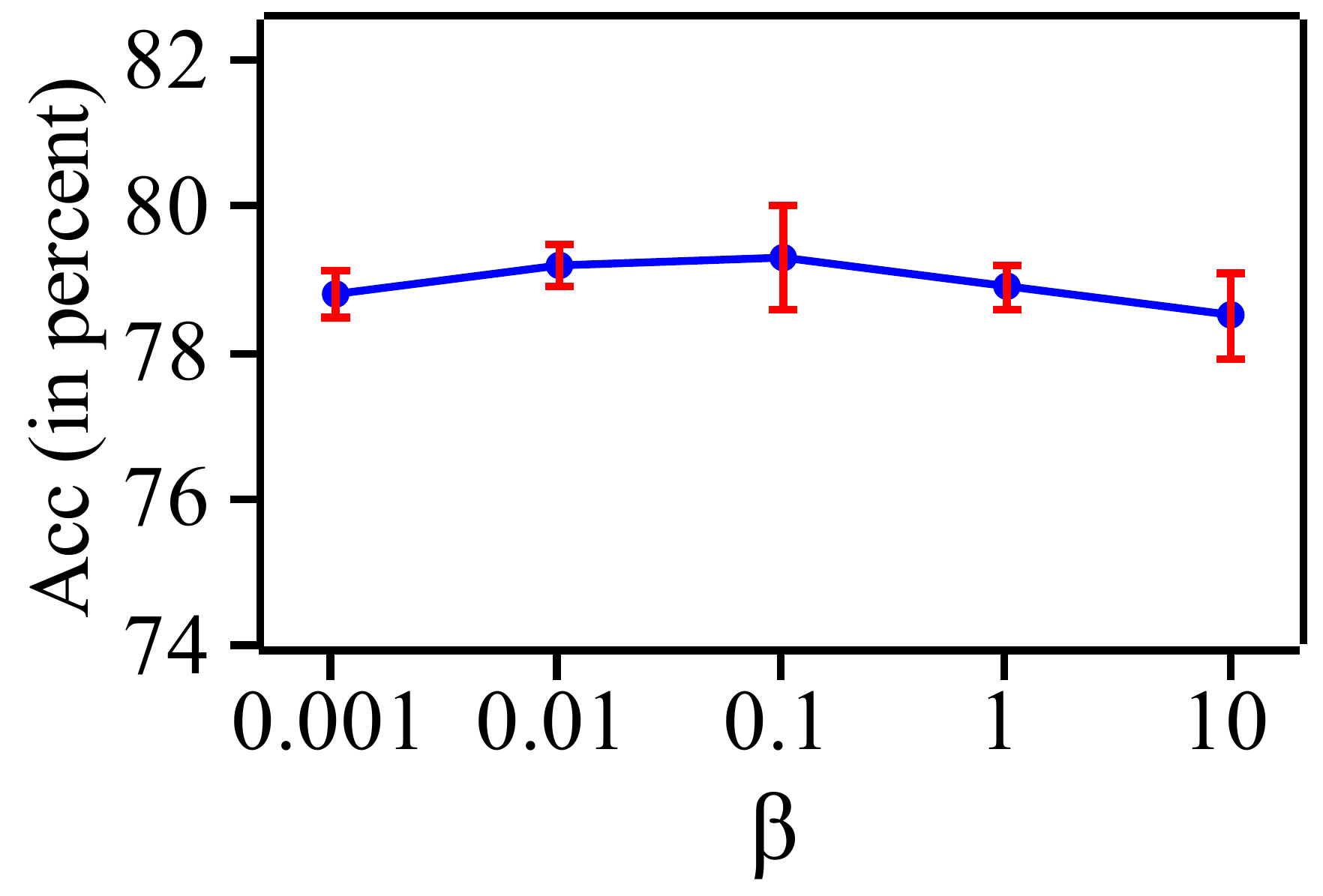}
		\end{minipage}%
	}%
	\subfigure{
		\begin{minipage}[t]{0.3\linewidth}
			\centering
			\includegraphics[width=0.95\linewidth]{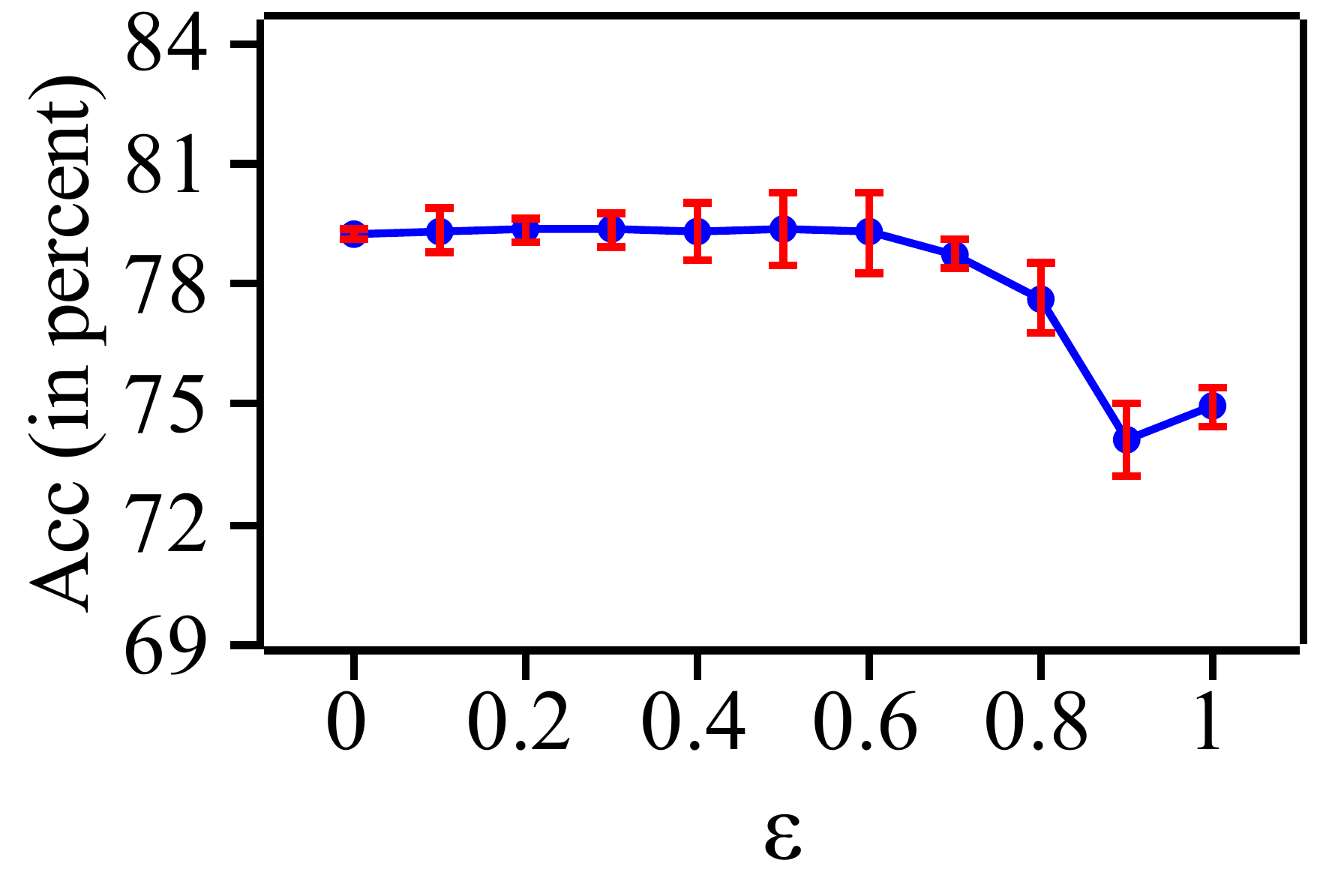}
		\end{minipage}
	}%
	\centering
	\caption{Classification accuracy by varying $\alpha$, $\beta$, and $\epsilon$, respectively, on clean Yummly10k dataset.}\label{fig:parameter}
\end{figure*}

When there are missing hyperedges or additional wrong hyperedges in the initial structure, the performance improvement of DeepHGSL will be more significant. Compared with HGNN$^{+}$, the improvement of classification accuracy of DeepHGSL is up to 16.6\% and 8\% with noisy hypergraph structure on Yummly10k and DBLP-coauthor datasets, respectively. Averaging over the four datasets, when 75\% of hyperedges are randomly deleted from the initial structure, DeepHGSL respectively achieves an average improvement of 8.8\%, 14.6\%, and 11.9\% compared with HGNN$^{+}$, HyperGCN, and HCHA.  And DeepHGSL boosts average accuracy over HGNN$^{+}$, HyperGCN, and HCHA by 7.33\%, 27.2\%, and 21.8\%, respectively, when 75\% of hyperedges are added. This means DeepHGSL can flexibly adjust the hypergraph structure according to the downstream task and make the network more robust to noise in the structure.  

\begin{figure*}[h]
	\centering
	\subfigure[Cora]{
			\includegraphics[width=0.23\linewidth]{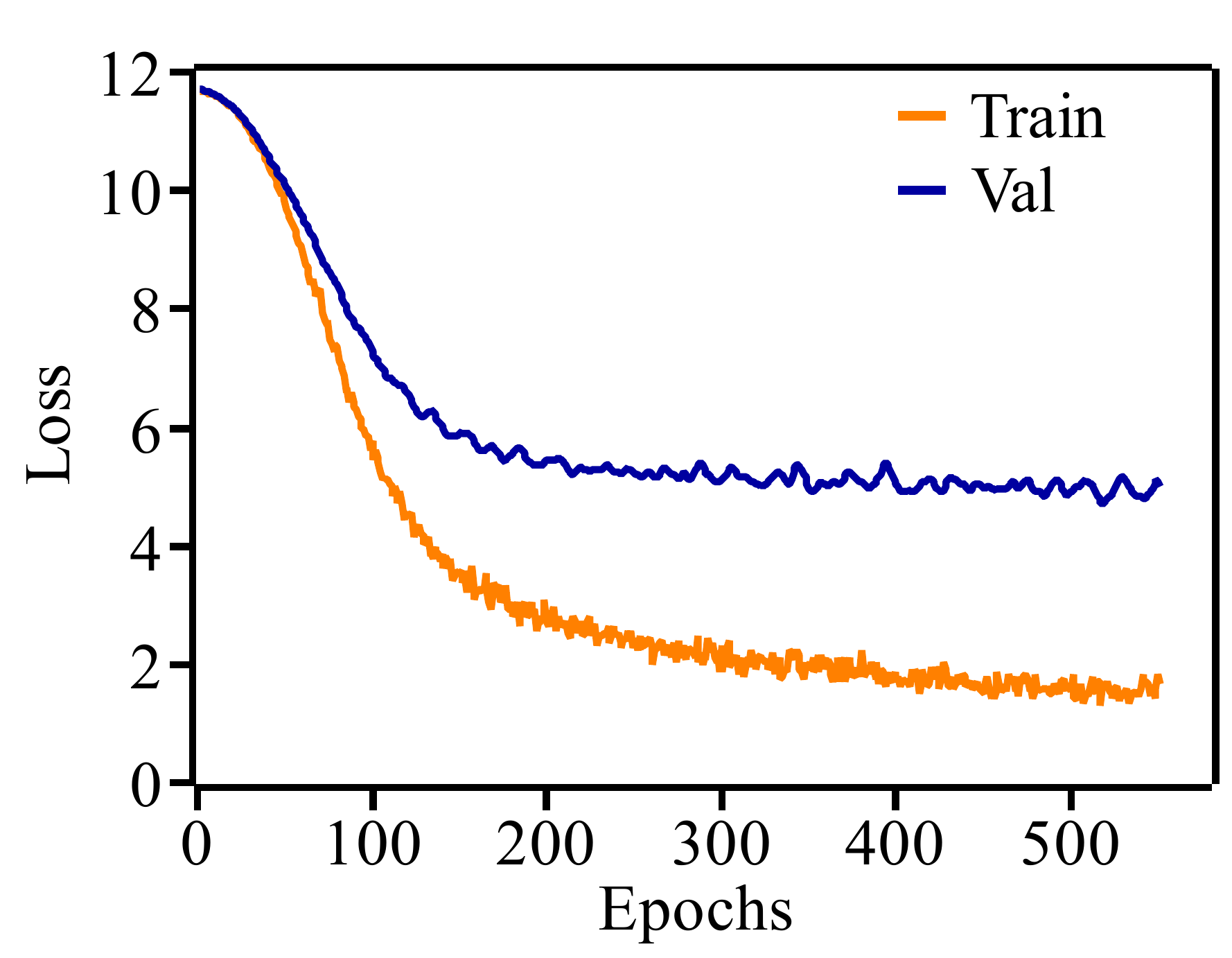}
	}
	\subfigure[Citeseer]{
			\includegraphics[width=0.23\linewidth]{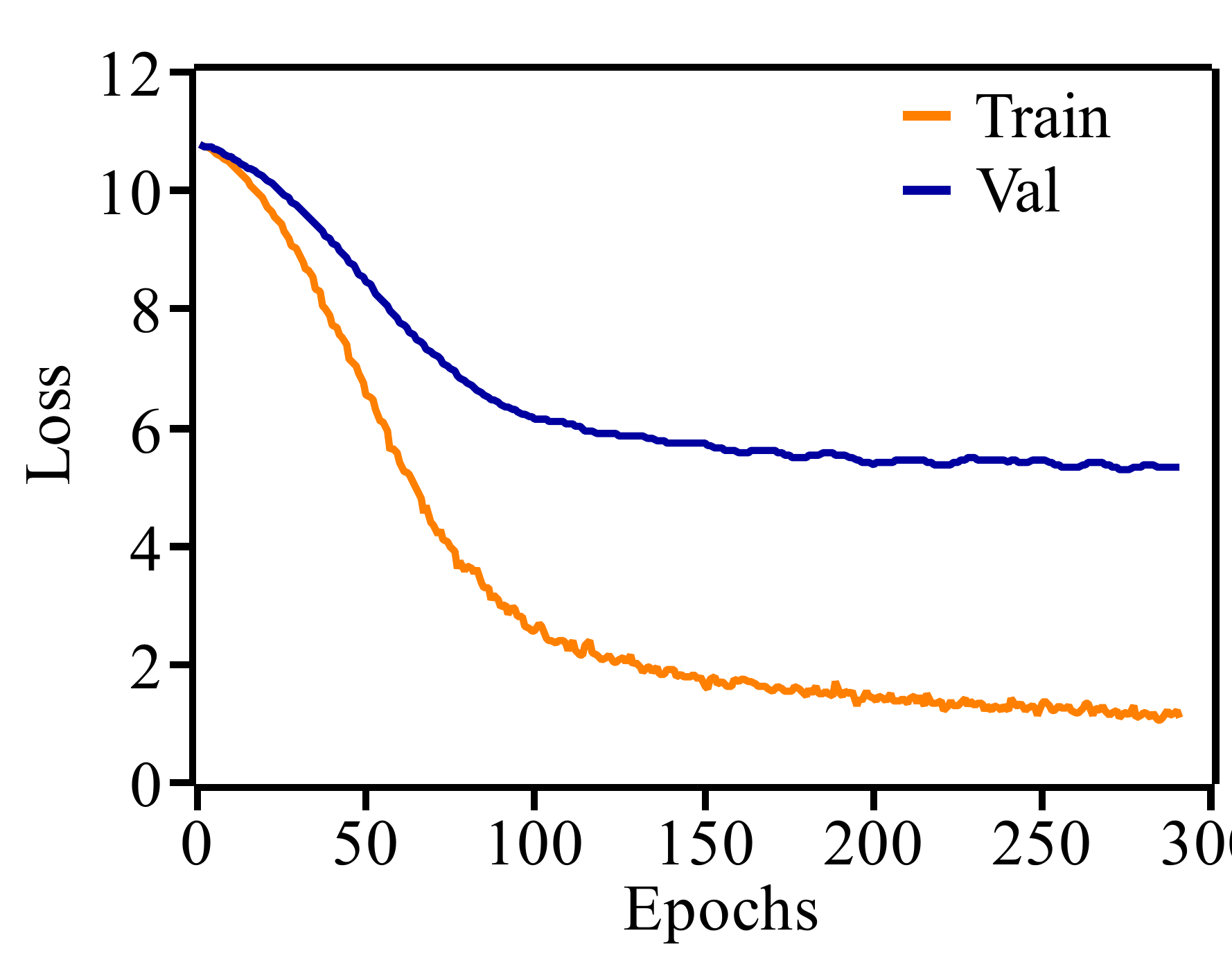}
	}
	\subfigure[Yummly10k]{
			\includegraphics[width=0.23\linewidth]{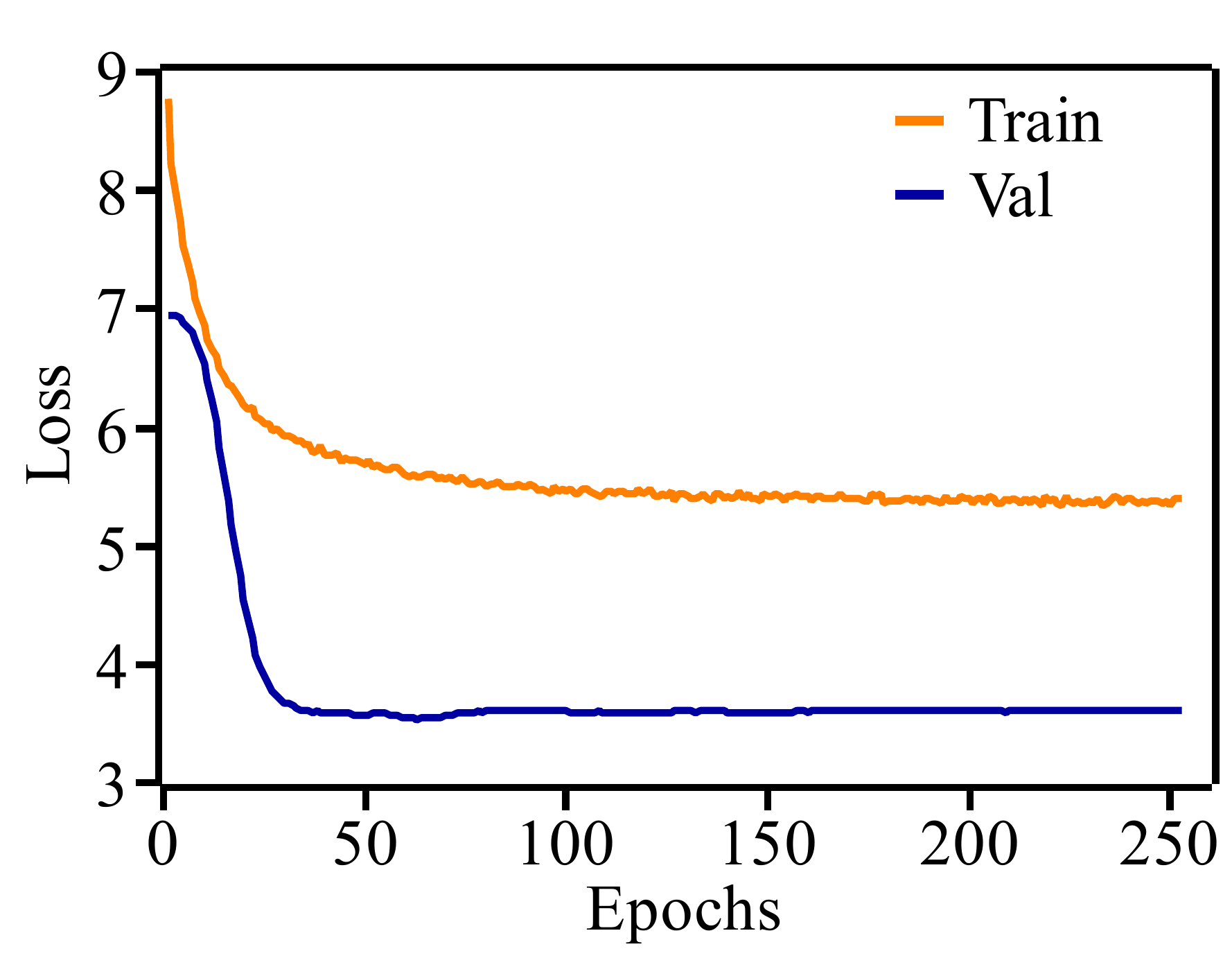}
    }
    \subfigure[DBLP-coauthor]{
			\includegraphics[width=0.23\linewidth]{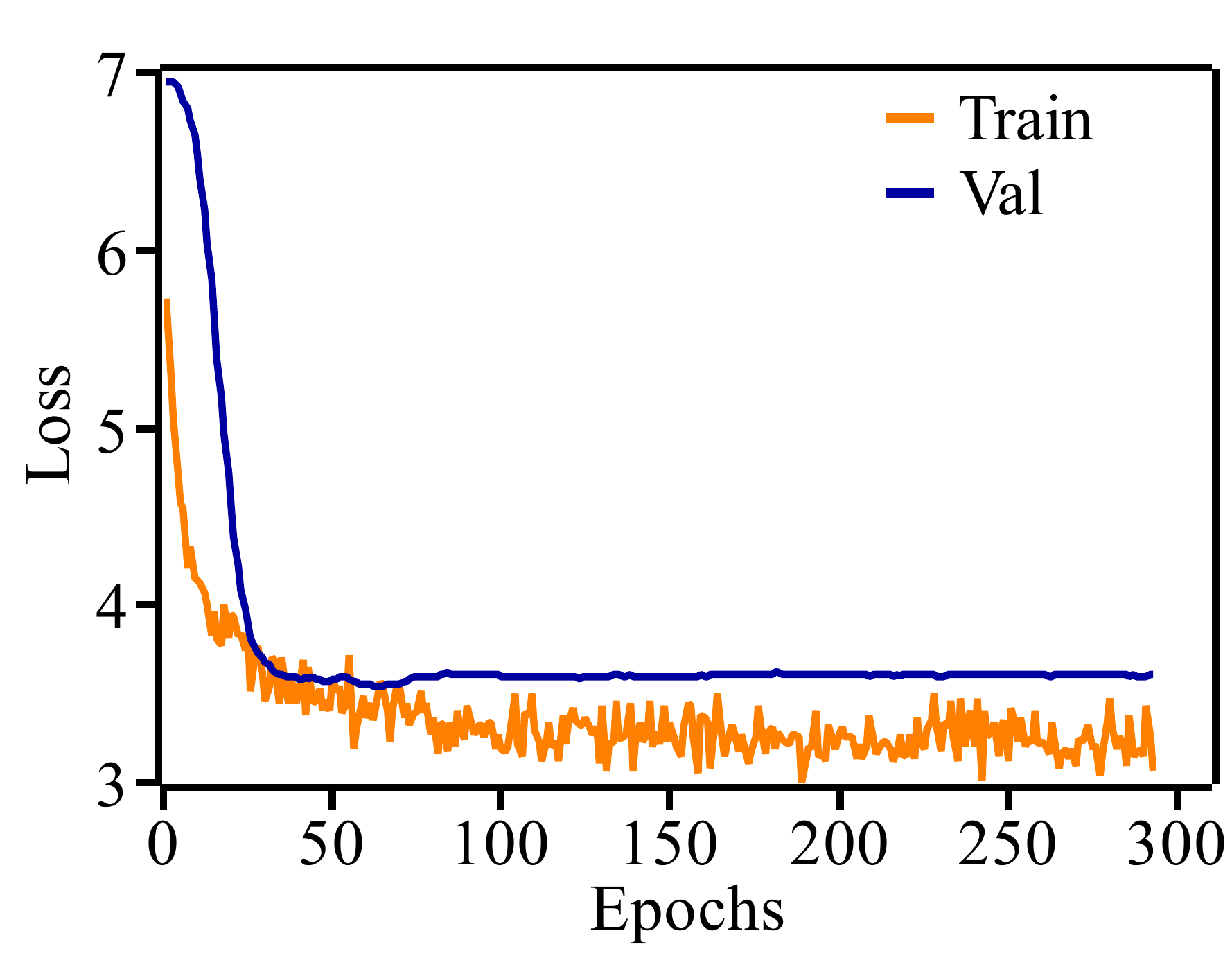}
    }
	\caption{Curves of training loss and validation loss on four datasets.}\label{fig:convergence}
\end{figure*}

On Yummly10k and DBLP-coauthor datasets, we further report the performance of a two-layer perceptron (MLP) only using node features, which has an average classification accuracy of 37.1\% and 71.8\%, respectively. An interesting observation is that the classification accuracy of MLP is much lower than graph/hypergraph based methods on Yummly10k dataset, and the accuracy of graph/hypergraph based methods drop sharply when the number of deleting hyperedges ascends, which means the high-order correlations among recipes based on their ingredients play an important role for predicting the cuisine labels. On the other hand, the classification accuracy of HGNN$^{+}$ is only 2.5\% higher than that of MLP on DBLP-coauthor dataset, indicating that the additional useful information in the coauthor relationship is limited for this task. This can explain the phenomenon that MLP even outperforms GAT and HyperGCN on this dataset by the fact that it is detrimental to embed too much useless information about the initial structure into the representation. The results demonstrate that our method can steadily improve performance regardless of whether the information in the initial structure is useful or redundant.

\subsection{Ablation Study}  

We conduct an ablation study on Yummly10k and DBLP-coauthor datasets to show the contribution of HIB loss to the performance, as shown in Fig. \ref{fig:abalation}. We use HIB to declare the model using the loss function in Eq. \ref{eq:loss}, and HIB-CE to denote the model only using cross-entropy as the loss function, which is written as
\begin{align}
\LM_{\text{HIB-CE}} & = \sum_{l=1}^{L}\LM_{\text{HIB-CE}}^{(l)} \notag \\
& = \sum_{v \in V}  CE \left(\hat{\y}_v^{(l)} ; Y_{v}\right).  \notag
\end{align}
The insight of $\LM_{\text{HIB-CE}}$ is to increase the mutual information between the representation and the target task without simultaneously reducing the mutual information related to the initial hypergraph structure. That is to say, the learned representations are likely to be affected by the noise information in the initial structure more severely. The experimental results demonstrate this point.
On average, HIB outperforms HIB-CE by 1.42\% and 1.01\% on Yummly10k and DBLP-coauthor datasets, respectively. This means that minimizing the mutual information between node embeddings and the initial structure can mitigate the effects of perturbations in the input data.

We also conduct experiments to see the effect of the number of layers in the framework on the classification performance. The curve in Fig. \ref{fig:abalation-layer} shows that updating the hypergraph structure at each layer contributes to the improvement of DeepHGSL. The classification accuracy is only 75.8\% when there is a single layer in the network, and rises up to 77.8\% when the network has multiple layers. When the number of network layers is increased, the performance generally shows an upward trend until stabilizing (\#Layers>7).

\subsection{Hyperparameter Sensitivity}

There are three essential hyperparameters $\alpha$, $\beta$, and $\epsilon$ in our model. $\alpha$ is used to control the weights of the initial hypergraph structure when updating the structure. $\beta$ is used to balance the impacts of two items in the HIB loss function. $\epsilon$ is the masking threshold of the attention scores to control the sparsity of the incidence matrix. Here, we investigate the influences of these three hyperparameters on classification accuracy. Every time we keep two  hyperparameters fixed at their best and change the third hyperparameter to see their influences on the results.  Experimental results on the clean Yummly10k dataset are demonstrated in Fig. \ref{fig:parameter}. In general, the performance is stable when these three hyperparameters change in a large range. We can see that a large $\alpha$ ($\alpha \leq 0.7$) may lead to relatively poor performance because too much information of the initial structure is retained in the process of updating the structure. When $\epsilon$ is too large, the performance also drops sharply  because of masking too many useful attention regions.

\begin{table}[]
	\centering
	\caption{Running time of the proposed method on all four datasets.} \label{tb:time}
	
	\begin{tabular}{lccc}
		\hline
		Dataset &  \#Parameters & Sec./epoch & Total Time (Sec.) \\ \hline
		Cora             & 31734        & 0.129  & 71.1    \\
		Citeseer        & 81658        & 0.362  & 105.3    \\
		Yummly10k         & 11504        & 0.161 & 40.7     \\ 
		DBLP-coauthor & 17072 & 0.205 & 60.1\\ \hline
	\end{tabular}
\end{table}

\subsection{Convergence Analysis and Running Time}
In our proposed DeepHGSL framework, the hypergraph structure is iteratively updated in each epoch. Therefore, the convergence speed of the learning process is important. Here, we study the convergence issue of our proposed method and draw the curves of training loss and validation loss on Cora, Citeseer, Yummly10k, and DBLP-coauthor datasets in Fig. \ref{fig:convergence}. With a step size of 0.01, our method can efficiently achieve convergence within 500 iterations on all four datasets. 
We conduct all experiments on a server with two Intel Xeon E5-2678
2.50 GHz CPUs and an Nvidia GeForce RTX 3090 GPU. The total number of parameters, mean running time per epoch, and total running time on four datasets are reported in Table  \ref{tb:time}. The total running time of the whole network is within two minutes, which is acceptable for practical applications. 

We note that our proposed method is a transductive learning method, and thus the operation is conducted on all nodes and hyperedges. High computational complexity is a common issue for transductive learning. For very large hypergraph datasets, it may raise out of memory issue. How to extend the deep hypergraph structure learning model and HIB principle to large-scale datasets is a future direction.
A feasible path to solve the structure learning problem of large-scale hypergraph-structured data is to select a few important  hyperedges and set them as anchors. The structure information about anchored hyperedges accounts for the majority of the total structure information. Only optimizing the high-order correlations related to the anchors will greatly reduce the computational cost.

\section{Conclusion}
In this work, we proposed a general paradigm, DeepHGSL, for deep hypergraph structure learning for robust node representation of hypergraph-structured data.  We introduced hypergraph information bottleneck to guide deep hypergraph structure learning and instantiated HIB as a trainable loss function. The minimal sufficient principle in HIB can reduce the noisy information in node embedding raised by the bad connections in the original hypergraph structure. We demonstrated the effectiveness and robustness of DeepHGSL in the face of 
different structural disturbances both on the graph-structured data and on the hypergraph-structured data. The experimental results validate that optimizing the hypergraph structure under the guidance of HIB can boost the performance of hypergraph neural networks. One of the future directions is to overcome the limitations caused by transductive learning and extend the current general paradigm to inductive learning to handle very large hypergraph datasets. What's more, an interesting problem is how to measure the useful information in the hypergraph structure for downstream tasks.  The mutual information between hypergraph structure and target output is a potential tool for the quantitative evaluation of hypergraph structure, which may promote the interpretability of hypergraph neural networks.

\ifCLASSOPTIONcaptionsoff
  \newpage
\fi



%
\bibliographystyle{IEEEtran}
\bibliography{ref}




\end{document}